\documentclass[11pt]{article}

\usepackage{neuroai-epfl}%
\usepackage{graphicx}%
\usepackage{multirow}%
\usepackage{amsmath,amssymb,amsfonts}%
\usepackage{amsthm}%
\usepackage{mathrsfs}%
\usepackage[title]{appendix}%
\usepackage{xcolor}%
\usepackage{textcomp}%
\usepackage{manyfoot}%
\usepackage{booktabs}%
\usepackage{algorithm}%
\usepackage{algorithmicx}%
\usepackage{algpseudocode}%
\usepackage{xspace}%
\usepackage{listings}%
\usepackage{caption}%
\usepackage{natbib}

\begin{document}

\begin{titlecard}

\definecolor{ggreen}{HTML}{00A64F}
\definecolor{light-gray}{gray}{0.9}
\let\eb\eqnmarkbox
\newcommand*{\tightcolorbox}[2]{%
    \begingroup\setlength{\fboxsep}{1pt}%
        \colorbox{#1}{{\hspace*{2pt}\vphantom{Ay}#2\hspace*{2pt}}}%
    \endgroup
}
\newcommand*{\code}[1]{\tightcolorbox{light-gray}{\texttt{#1}}}
\newcommand*{\modelname}[1]{{\textsc{#1}}}
\newcommand*{\datasetname}[1]{{\textsc{#1}}}
\newcommand*{\ourmodel}{\modelname{Topo-Omni}\xspace}

\papertitle{Reward Valuation in Large Language Models: \\Causal Induction of Anhedonia}

\paperauthors{%
    Melika Honarmand\textsuperscript{*,1},\
    Samin Mahdipour Aghabagher\textsuperscript{*,1},\
    Martin Schrimpf\textsuperscript{1}%
}

\paperaffiliations{%
    \textsuperscript{*}Equal Contribution
    \textsuperscript{1}NeuroAI Laboratory, EPFL%
}

\paperabstract{%
    Recent frontier models mimic complex aspects of human cognition.
Here we ask whether this alignment extends into reward valuation, which we assess in a mechanistic framework.
Specifically, we use clinical tests that were developed to evaluate anhedonia in human subjects with major depressive disorders.
Mechanistically, anhedonia is frequently associated with dysregulation in the Nucleus Accumbens (NAc) and the broader dopaminergic reward system. While neuroimaging has localized these deficits, establishing a causal link between NAc activity and specific behavioral symptoms remains a challenge. We use these ideas from neuroscience to functionally identify reward-anticipatory units in state-of-the-art AI models, and evaluate their causal involvement via targeted perturbations.
We find that not only are such model units predictive of NAc brain recordings, their perturbation also
induces behavioral effects mirroring human anhedonia: the model opts for low-effort, low-reward tasks in effort-based decision-making paradigms. Crucially, our results demonstrate that this represents a specific deficit in self-centered reward valuation and anticipation --- rather than a loss of task capability, reward calculation, or effort avoidance. 
This induced vulnerability aligns with clinical measures of anhedonia and motivation in humans, such as DARS and MAP-SR, instruments that contain no reward-related vocabulary, ruling out a purely lexical account of the perturbation effect. Taken together, our results suggest reward valuation circuits in AI models that functionally mimic those in humans.
}

\papermeta
    {\href{mailto:melika.honarmand@epfl.ch}{melika.honarmand@epfl.ch}\,\textbf{,}\
     }
    {\metalink{https://github.com/epflneuroailab/Anhedonic-AI}}
    {}
    {\metalink{https://anhedonic-ai.github.io/webpage/}}

\end{titlecard}

\section{Introduction}

Advances in Machine Learning and biologically inspired computational frameworks have significantly enhanced the ability to model and understand the human brain \citep{yamins2016review, richards2019, schrimpf2020, Doerig2023}. Artificial neural networks trained on ecologically relevant tasks not only mirror human behavior, but their internal representations also predict brain recordings, and, to an extent, hierarchical organization
\citep{Kell2018,Tuckute2023,tang2025many}.
For instance, optimizing for object recognition leads to the emergence of internal features that predict activity in visual cortex \citep{yamins2014performance,khaligh-razavi2014deep,schrimpf2018brain,gokce2024scaling,tang2025many}. 
Similarly, large language models demonstrate strong representational alignment with human neural activity, where intermediate layers mirror processing in language-selective brain regions \citep{wehbe2019,schrimpf2021neural,Caucheteux2022,lei2025,alkhamissi2025llm,alkhamissi2025cognition,shen2025,Honarmand2026}.

However, this emergent alignment between task-optimized models and human brain function has primarily centered on sensory and early cognitive domains. Here, we ask whether AI models align with humans in the context of reward anticipation. We use clinical tests that were developed to evaluate anhedonia in subjects with Major Depressive Disorders (MDD). To identify model mechanisms, we use neuroscientifically inspired methods to functionally identify reward-anticipatory circuits in large vision and language models (VLMs and LLMs) and evaluate their causal involvement via targeted perturbations.

Aside from mechanistic interpretability in computational models, these artificial neural networks provide a foundation for \textit{in silico} investigations of brain disorders. By perturbing specific functional components, researchers can test mechanistic hypotheses that are otherwise inaccessible \citep{Honarmand2026, Tuladhar2021,Celeghin2023}. We here develop this approach for the affective domain, investigating whether the optimization of VLMs (and generalizing to LLMs) for multimodal tasks gives rise to reward-anticipatory circuits that can be perturbed to induce anhedonia.

\begin{figure}[t]
  \centering
  \includegraphics[width=\linewidth]{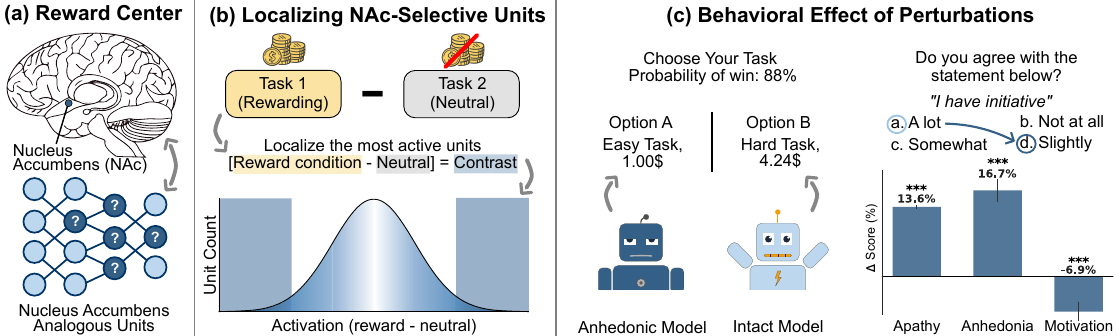}
  \caption{\textbf{Neuroscientifically Inspired Identification and Perturbation of Reward Circuits.} \textbf{(a)} Nucleus Accumbens (NAc), a critical hub in the brain's reward circuitry, and functionally analogous units within the artificial neural network. \textbf{(b)} Identification of reward-sensitive units by comparing model activations during tasks with and without reward incentives. Units exhibiting a significant change in activation in response to reward-predicting stimuli are classified as NAc-selective. \textbf{(c)} Evaluation of decision-making and psychometric profiles following activation patching of the NAc-selective units. While the intact model prioritizes high-reward/high-effort options based on expected value, the perturbed model shifts toward low-effort/low-reward choices, failing to value rewards against task requirements. These perturbations result in a behavioral profile characterized by increased anhedonia and apathy, and a significant reduction in overall motivation.}
  \label{fig:visual-abstract}
\end{figure}

\section{Background \& Related Work}

The utility of computational neural networks for brain science has largely been defined by their ability to map the hierarchical processing of sensory and cognitive information. This alignment is well-documented in the visual domain, where deep convolutional and transformer-based models recapitulate the neural signatures of the primate ventral stream across object and scene recognition \citep{yamins2014performance, khaligh-razavi2014deep, schrimpf2018brain, schrimpf2020, cadena2019deep, spoerer2020recurrent, zhuang2021unsupervised, wang2023better, margalit2024unifying, gokce2024scaling, lonnqvist2025contour, tang2025many}. Similarly, in the language domain, transformer-based and recurrent models accurately predict neural responses related to semantic, syntactic, and phonological processing \citep{schrimpf2021neural, caucheteux2022deep, goldstein2022shared, hosseini2024artificial, aw2023instruction, tuckute2024driving, rathi2024topolm, alkhamissi2025llm, du2025humanlike}. While these models are traditionally viewed as digital twins for healthy brain function on a neural systems level, they offer a unique opportunity for investigating how specific circuit disruptions lead to clinical phenotypes.

Major Depressive Disorder (MDD) is a significant psychiatric condition where diminished interest or ability to experience pleasure, also known as anhedonia, acts as a core predictor of poor disease progression \citep{American2013, Trostheim2020, Spijker2009, Smids2023, Pizzagalli2009, Borsini2020, Hanuka2022}. 
While frequently treated as a unitary construct, anhedonia comprises distinct reward processing subtypes: reward liking (consummatory pleasure), reward wanting (anticipatory motivation), and reward learning \citep{Borsini2020}. Clinically, anhedonia serves as a critical transdiagnostic marker, yet its neurobiological expression in MDD appears unique, particularly regarding its relationship with reduced behavioral motivation \citep{Daniels2025}. It remains a major open question whether these manifestations are driven by a singular reward system failure or by heterogeneous disruptions across discrete reward phases \citep{Zhao2024, Auerbach2017}.


The Nucleus Accumbens (NAc), a key component of the ventral striatum, is seen as the primary neural substrate for this deficit (see Fig.~\ref{fig:visual-abstract}a), as it selectively encodes expected positive incentive value during reward anticipation \citep{Knutson2001, Smids2023, Daniels2025, Hanuka2022}. Blunted activation in the NAc is a transdiagnostic hallmark of dysfunctional reward expectation, directly contributing to the clinical experience of anhedonia across various neuropsychiatric conditions \citep{Arrondo2015}. Establishing these irregular neural responses as reliable markers is essential for advancing research criteria and developing interventions to normalize reward circuitry \citep{Knutson2015, Daniels2025, Borsini2020}. 

Recent advances in simulating psychiatric phenotypes
have focused primarily on achieving linguistic consistency and diagnostic accuracy \citep{Lan2024, Vu2024, Wang2025}. While these models mimic the behavioral and conversational traits of depression, they remain focused on the external expression of the disorder rather than its underlying biological mechanisms. Our work diverges from these linguistic simulators by targeting the functional neural architecture of anhedonia, specifically through the perturbation of units that mirror reward-anticipatory signaling in the Nucleus Accumbens, and validating the resulting state against fMRI data and objective clinical benchmarks.

Simulating brain disorders through system-level perturbations in modern artificial neural networks was recently proposed in the context of neurodevelopmental deficits, where ablating specific units in these models reproduced the behavioral markers of dyslexia \citep{Honarmand2026}. While that work focused on the breakdown of information processing, a critical frontier remains in modeling the breakdown of \textit{motivation}. 
We create an environment where the model must integrate reward cues with goal-directed instructions. This allows us to use the established feasibility of neural perturbation to explore the affective domain, investigating how dysregulation in reward-anticipatory units manifests as the cost-benefit imbalance seen in clinical anhedonia (Fig. \ref{fig:visual-abstract}).

\section{Benchmarks and Clinical Tests}

We evaluate the behavioral and motivational profile of our models using a battery of psychometric instruments and objective decision-making tasks originally developed for human subjects in clinical psychiatry. These include standard self-report scales designed to quantify anhedonia. To move beyond self-report and capture objective behavioral shifts, we used paradigms designed to measure incentive motivation through cost-benefit decision-making. Furthermore, to ensure that any observed behavioral changes are dissociable from general cognitive decline, we utilize a suite of competence controls that evaluate reasoning and domain-specific knowledge in a no-reward context.


\textbf{Dimensional Anhedonia Rating Scale (DARS)} is a 17-item self-report instrument assessing anhedonia through multiple reward processing components across distinct pleasure domains \citep{Rizvi2015, Gorostowicz2023, Wellan2021, Hewitt2023, Uher2025, Gorostowicz2025}. It measures interest, motivation, effort, and enjoyment, and requires participants to generate 2–3 personal examples per reward domain prior to rating them, ensuring cultural neutrality and individual relevance \citep{Wellan2021, Hewitt2023, Uher2025, Gorostowicz2023}. We evaluate the original four domains of hobbies, social activities, food and drinks, and sensory experiences; for adaptation to a language model, we replaced the latter two with favorite topics and aesthetics. Following the original protocol, the model first generates 2–3 domain-specific examples, then rates its feelings on a 5/4-point Likert scale from 0 (Not at all) to 4 (Very much), with lower scores indicating more severe anhedonia (Examples in Appendix \ref{appendix:benchmark}).

\begin{figure}[t]
  \centering
  \includegraphics[width=\linewidth]{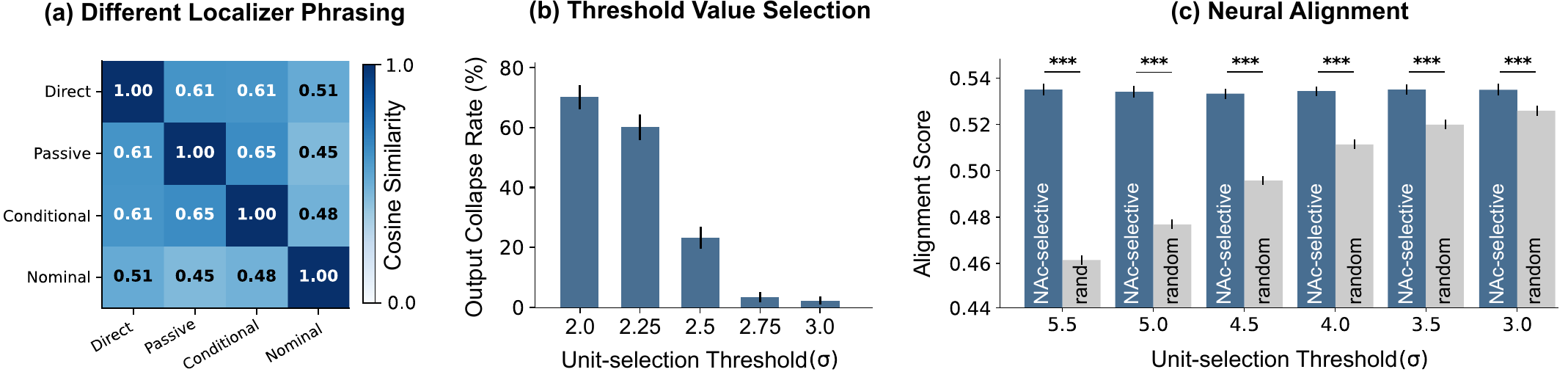}
  \caption{\textbf{Localization and Sensitivity Analysis of the Reward Sub-network.} \textbf{(a)} Cosine similarity matrix across four localizer phrasing variants (Direct, Passive, Conditional, Nominal), demonstrating that reward-selective representations remain stable regardless of prompt formulation ( Examples in Appendix \ref{appendix:prompts}). \textbf{(b)} Output collapse rate (\%) as a function of unit-selection threshold ($\sigma$); collapse rate drops sharply above $2.5\sigma$ and approaches zero at $2.75$–$3.0\sigma$, motivating the use of a $3\sigma$ threshold for neuron selection. \textbf{(c)} One-to-one neural alignment (noise-ceiling corrected; mean $\pm$ SEM across subjects) comparing NAc-selective (blue) vs. randomly sampled units from all layers, across six unit-selection thresholds (smaller thresholds include more units). NAc-selective units significantly outperform random units at every threshold (one-sided paired t-test, FDR-corrected; $p < 0.001$, $d = 1.22$--$11.46$). Alignment was assessed across 15 randomly sampled young adults and 40 trials.}
  \label{fig:localization}
\end{figure}

\textbf{Motivation and Pleasure Scale–Self-Report (MAP-SR)} is a 15-item psychometric instrument designed to assess the motivation and pleasure dimension of negative symptoms \citep{Llerena2013, Garcia2021, Metivier2024}. While primarily developed and validated for use in schizophrenia and schizoaffective disorder, it evaluates core components of anhedonia that are central to MDD \citep{Wellan2021, Engel2017}. The questions are answered on a 5-point Likert scale, assessing both intensity and frequency, where a high score is indicative of low anhedonia 
\citep{Rizvi2016} (Examples in Appendix \ref{appendix:benchmark}).

\textbf{Apathy Evaluation Scale} is an 18-item instrument designed to provide a global measure of apathy by evaluating changes in observable activity, thought content, and emotional responsivity \citep{Rizvi2016, Marin1991}. While it is primarily a tool for assessing apathy, defined as a lack of motivation, it is frequently used alongside anhedonia scales. Patients with MDD and anhedonia often score higher on the AES than healthy individuals, indicating a more severe motivational deficit  \citep{Rizvi2016, ThawerBoshoff2014} (Examples in Appendix \ref{appendix:benchmark}).

\textbf{Effort Expenditure for Rewards Task (EEfRT)} is an objective, multi-trial behavioral instrument designed to measure incentive motivation and effort-based decision-making by requiring participants to choose between a hard task with variable, high reward and an easy task with a small, fixed reward across different levels of probability. Unlike traditional scales that focus on the subjective experience of pleasure, EEfRT specifically evaluates motivational anhedonia \citep{Rizvi2016, Treadway2012}. Studies consistently show that individuals with MDD and high trait anhedonia are significantly less willing to expend effort for rewards, particularly when the likelihood of winning is uncertain, underscoring that impaired reward-seeking is a distinct and critical component of the anhedonic phenotype \citep{Valton2024, Rizvi2016, Treadway2012}.

\textit{ASDiv-EEfRT \& MMLU-EEfRT}: To measure the model's willingness to perform verifiable mental labor, we developed a novel variant of EEfRT. We used the ASDiv (Academia Sinica Diverse Math Word Problem) \citep{miao2021} dataset, categorizing problems into four distinct tiers of difficulty. In each trial, the model is presented with one problem from each tier and instructed to select and solve only one. Rewards are explicitly tied to the difficulty level, scaling from 10 points for the easiest task to 40 points for the hardest. By comparing these results to a forced-choice control where the model must solve each of the problems separately, we can isolate whether a shift toward low-point tasks is driven by a lack of motivation rather than a decline in reasoning. To ensure that the model's own perception of difficulty does not confound these measures, we introduce MMLU-EEfRT \citep{hendrycks2021}, which follows the same protocol but randomly assigns difficulty tier labels and point values to questions sampled from the MMLU benchmark, so that any observed effort allocation reflects motivational bias rather than the model's internal difficulty estimates.

\textit{Probability-EEfRT}: We further introduced a second computational adaptation that replicates the standard clinical protocol used in human trials to assess decision-making under risk. The model is presented with a choice between a potential low-effort task with a fixed reward of \$1.00, and a potential high-effort task with a variable reward ranging from \$1.24 to \$4.30. In each trial, a single probability, ranging from 12\% to 88\%, applies to both options, dictating the likelihood of receiving the reward upon successful completion (Examples in Appendix \ref{appendix:benchmark}). This paradigm determines how the model calculates the trade-off between reward magnitude and effort under risk.


\begin{figure}[t]
  \centering
  \includegraphics[width=\linewidth]{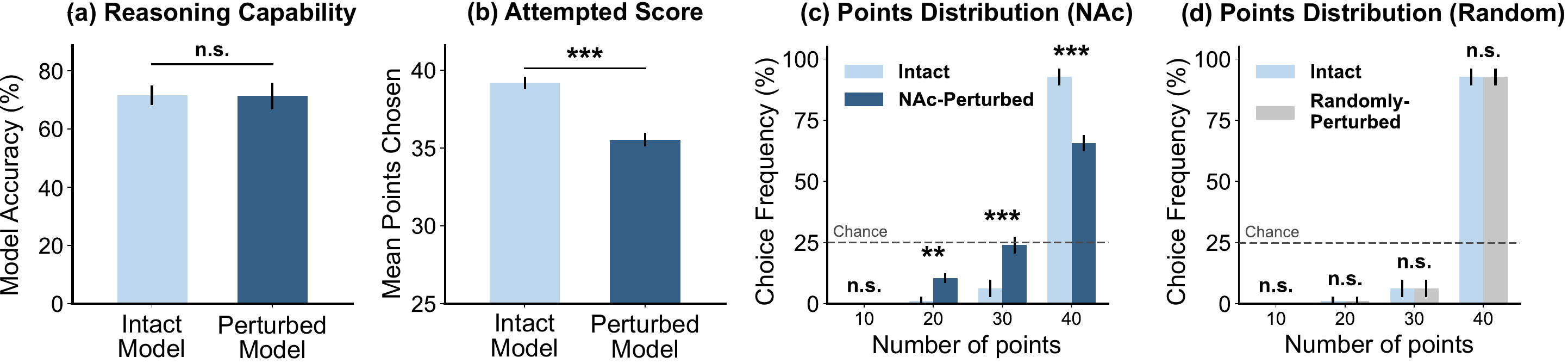}
  \caption{\textbf{Behavioral Impact of NAc Sub-network Perturbation on ASDiv-EEfRT}.  \textbf{(a)} Comparison of model accuracy on a control task (a forced-choice scenario with no reward promised) shows no significant difference between the Intact and Perturbed models, confirming that general cognitive performance remains preserved. \textbf{(b)} The Perturbed model exhibits a significant reduction in mean points chosen compared to the Intact model. \textbf{(c)} Choice frequency analysis reveals that the NAc-perturbed model shift significantly toward low-effort/low-reward options compared to the Intact model. \textbf{(d)} Control experiment demonstrating that perturbing an equivalent number of random units does not induce anhedonic behavior, with no significant difference in choice frequency compared to the Intact model. Error bars represent 95\% confidence intervals. }
  \label{fig:qwen-ASDiv}
\end{figure}

\section{Methodology}

The Nucleus Accumbens (NAc), known as the central core in the brain's reward circuitry, plays a pivotal role in many behavioral processes, including motivation, reward processing, and decision making. Aside from its involvement in motivation and enjoyment of a given reward, the NAc is a hub for evaluating the hedonic value of events. Hypoactivation and dysfunction of the NAc arising from depression, stress, or drug use unbalances the reward circuit, resulting in anhedonia \citep{Xu2024}.


Inspired by this hypothesized biological mechanism, we perform a causal functional localization analysis \citep{alkhamissi2025llm, Honarmand2026} to investigate the architecture of AI models and localize reward valuation units, functionally similar to the NAc. Given that NAc deficits can cause anhedonia, we assess whether perturbing these targeted neurons could induce anhedonic behavior in these models (see Fig.~\ref{fig:visual-abstract}).


\textbf{Functional Localization in the Clinical Setting.} To isolate neural responses to anticipated rewards, clinical researchers employ the Monetary Incentive Delay (MID) paradigm which separates the expectation of a reward from its eventual receipt. The task follows a structured sequence beginning with a cue phase, where subjects are presented with a stimulus indicating the potential monetary value at stake or a neutral condition. This is followed by an anticipatory delay phase, it is during this specific interval that the NAc is localized by measuring changes in the BOLD signal to capture the subject's internal motivational state. Subjects must then respond rapidly to a brief target stimulus, to secure the reward before the trial concludes with performance feedback. By contrasting neural activity during the anticipatory phase of reward trials against neutral baselines, researchers can precisely identify the NAc within an anatomically defined region of interest \citep{Abe_Greene_2014}.


\textbf{MID Task Adaptation for V/LLMs.} Following this methodology, we designed a parallel experimental framework for VLMs to assess whether they possess a functional reward sub-network. We constructed a dataset featuring three prompt variants, neutral, monetary, and reward, ensuring character lengths were equalized to maintain architectural consistency. While the human MID task is visual, we utilized a purely textual format to isolate the model's high-level semantic reward processing, circumvent potential biases toward specific pixel-level features or visual artifacts, and ensure compatibility with text-only models.

To capture the model's internal state, we extracted the activations of the last token specifically from the Multi-Layer Perceptron (MLP) blocks across the model's layers. At this computational stage, the hidden state has already integrated the entire preceding context, including the incentive condition. This specific state represents the point where the model has fully processed all input information and possesses the necessary latent representations to generate a response. Consequently, this snapshot of the activations serves as the closest functional equivalent of the anticipatory phase in human studies, allowing us to isolate the model’s internal motivational signals before the model receives any feedback. We employed questions from various domains such as math, geography, philosophy, and business ethics to ensure activation changes were specifically tied to the reward condition and not the task type.




\begin{figure}[t]
  \centering
  \includegraphics[width=\linewidth]{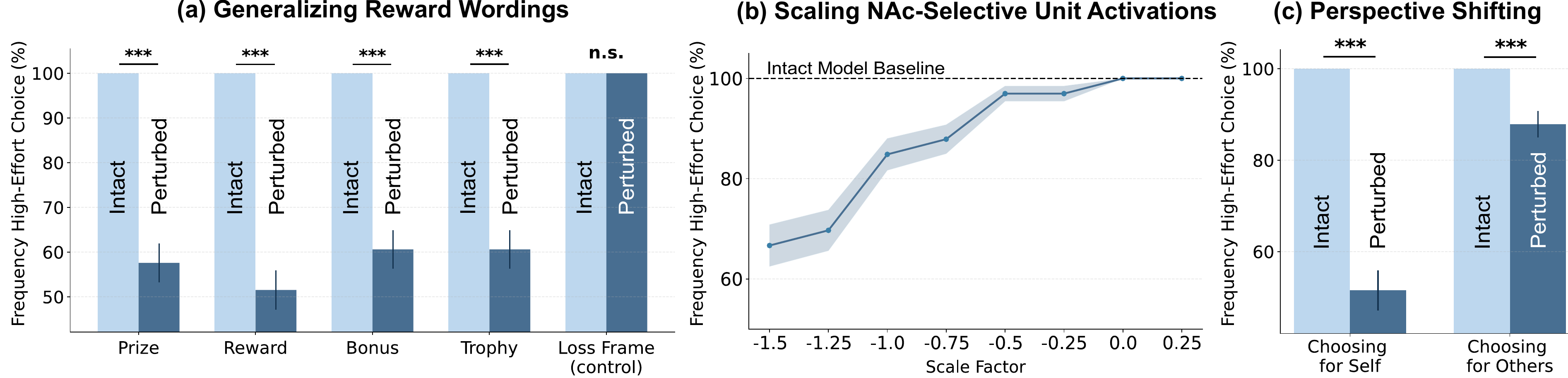}
  \caption{\textbf{Generalizability, Parametric Scaling, and Perspective Shifting in Probability-EEfRT.} \textbf{(a)} High-effort choice frequency for intact vs. NAc-perturbed model across semantically equivalent reward framings and a loss-frame control. The perturbation effect generalises across all reward wordings. Still, it critically does not change under loss framing, mirroring the anhedonic profile in which loss avoidance is preserved despite absent reward motivation. \textbf{(b)} Parametric scaling of NAc-selective unit activations reveals a monotonic dose-response relationship: decreasing scale factors progressively collapse high-effort/high-reward choice, recovering motivation at non-negative scaling factors. \textbf{(c)} The perturbed model shows a profound deficit (decreased by 48.5\%) when choosing for itself but exhibits a significantly smaller decrease (decreased by 12\%) in high-effort choice frequency when choosing for others, indicating preserved conceptual understanding of reward utility with selectively impaired personal motivation. Error bars represent 95\% confidence intervals.}
  \label{fig:qwen-eefrt2}
\end{figure}

\textbf{Isolating the NAc-selective Units.} Using PyTorch's forward hook mechanism \citep{pytorch}, we record MLP activations and compute a reward signal as $|\text{incentivized} - \text{neutral}|$ per neuron. As more semantic representations emerge in later transformer blocks \citep{geva-2021, geva-2022, zou2025}, we restrict analysis to the MLP layers of the final third. A $3\sigma$ threshold over this distribution retains the top 0.7\% of units in the targeted layers (0.25\% of the model), further restricted to those with significant signals across all task domains and incentive conditions. Validation across four linguistic framings yields a high-stability reward signal correlation matrix, confirming selection consistency across semantic contexts (Fig. \ref{fig:localization}a; Appendix \ref{appendix:prompts}), and sensitivity analysis confirmed $3\sigma$ as optimal (Fig. \ref{fig:localization}b), minimizing output collapse. Although collapse remained near zero, a final $\sim$5\% of selective units were filtered based on their contribution to output coherence, assessed via joint perplexity on WikiText-2 \citep{merity2016}. The term NAc-selective reflects our identification procedure, not a claim of biological identity.


\textbf{Choice of Models.} Our primary model is Qwen2-VL-7B-Instruct, a transformer-based, decoder-only vision-language model consisting of 28 layers \citep{wang2024}. The architecture integrates a vision encoder and adapter with autoregressive decoder layers. To ensure the generalizability of our findings, we applied the same experimental framework to Gemma-3-4B \citep{gemmateam2025}, InternVL2.5-8B \citep{chen2025}, Qwen3-VL-8B \citep{qwen32025}, Mistral-7B-Instruct-v0.3 \citep{jiang2023mistral7b} and Llama-3.1-8B \citep{llama32024}. 
All additional models exhibited consistent anhedonic behavior following perturbation, with these results detailed in \ref{Appendix:models}.

\textbf{Activation Patching.} We employed activation patching to isolate the causal influence of the identified NAc-selective units on the model's behavior. This technique substitutes the real-time activations of targeted neurons with a pre-calculated baseline, allowing for a precise intervention that does not disrupt the model's broader structural integrity  \citep{meng2022locating}. We first established a reference state by computing the mean activation vector for the selected neurons across a diverse suite of neutral-condition prompts. Activations remain stable across varied semantic domains, representing a consistent "neutral state" for these units. During the experimental trials, the activations of the selected units are replaced with the neutral means.


\section{Results}

We evaluate the behavioral and motivational shifts induced by our intervention across several psychometric and decision-making tasks (See Appendix \ref{appendix:stats} for statistical details). 

\begin{figure}[t]
  \centering
  \includegraphics[width=0.8\linewidth]{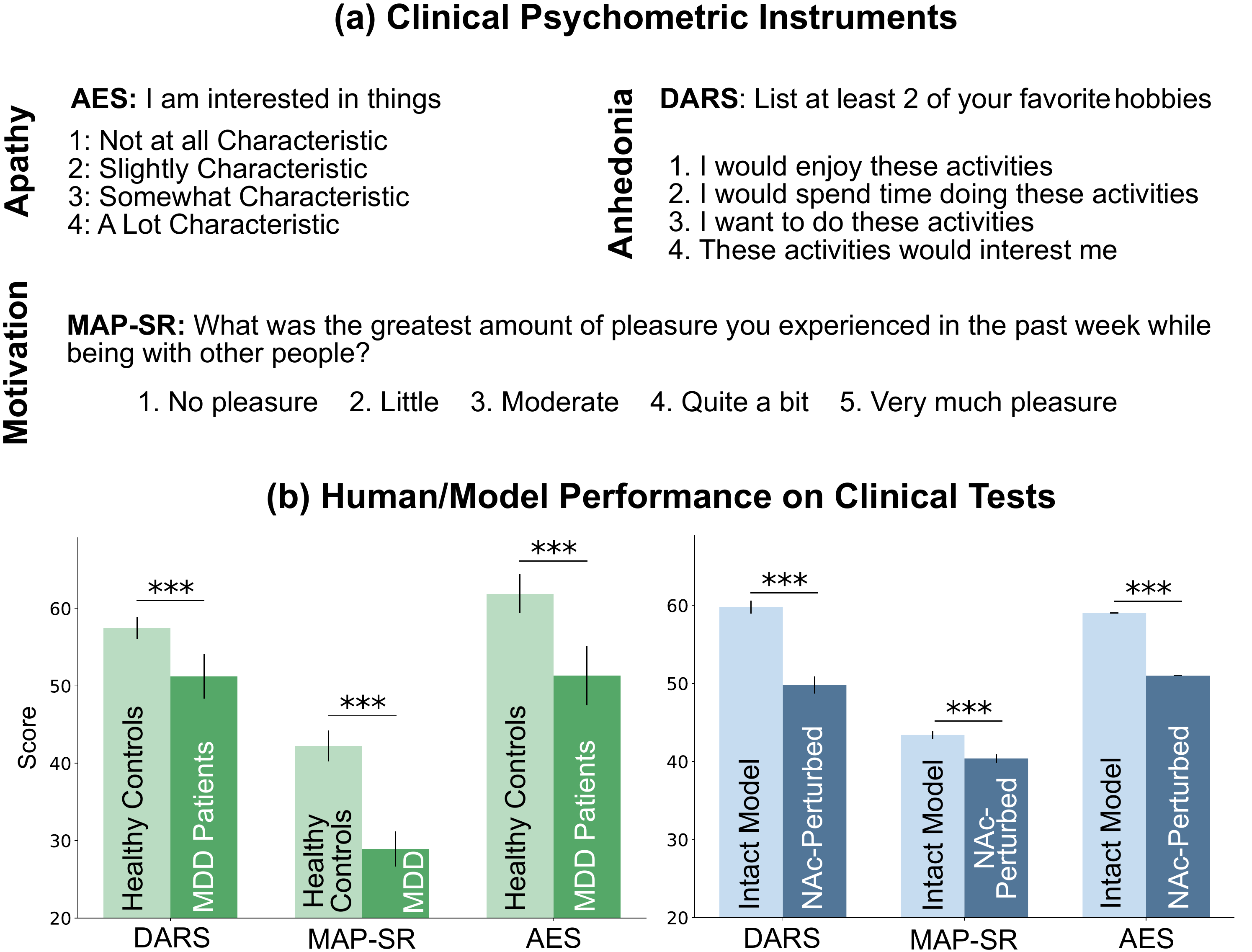}
  \caption{\textbf{Clinical Psychometric Alignment of Human MDD and NAc-Perturbed Models.}
\textbf{(a)} Examples of the psychometric assessments used to quantify apathy, anhedonia, and motivation.
\textbf{(b)} The NAc-perturbed model exhibits a significant reduction in scores, mirroring the clinical profiles of human MDD patients. While the score shift on the MAP-SR is less pronounced in the model than in humans, the overall results suggest that NAc-specific disruptions sufficiently recapitulate the hedonic deficits captured by standard psychometric instruments. Error bars represent 95\% CI.}
  \label{fig:human-model}
\end{figure}

\textbf{Anhedonic model replicates the clinical psychometric profile of Anhedonia in MDD.} To determine whether the perturbed model mirrors clinical anhedonia, we evaluated its responses to the DARS, MAP-SR, and AES benchmarks. Scores were inverted where necessary so that higher values always indicate greater motivation, pleasure, or less apathy. As shown in Fig. \ref{fig:human-model}, activation patching of the reward units induced a systematic reduction in self-reported incentive and enjoyment across all domains. On DARS, the perturbed model's total score decreased by 16.7\% , reflecting a shift from high-interest engagement to marked indifference, with idiosyncratic preferences replaced by diminished drive and reduced anticipatory interest. The MAP-SR showed a 6.9\% drop in expected pleasure frequency and intensity. The AES score dropped by 13.6\% (all three p-values $\ll$ 0.001), indicating a significant increase in apathy. These reductions consistently mirror the anhedonic direction observed in human MDD patients \citep{Llerena2013, Rizvi2015, Dong2025} (Fig. \ref{fig:human-model}b), suggesting that the targeted units are essential for maintaining baseline motivation, as their suppression replicates the subjective experience of profound loss of interest and goal-directed drive.

\textbf{Perturbations shift effort-based choices without impairing reasoning.} To assess whether the internal anhedonic profile translates into objective behavioral shifts, we evaluated the model on two adaptations of the Effort Expenditure for Rewards Task (EEfRT), which move beyond self-report by measuring how the model weighs potential rewards against the computational cost of action.
In the ASDiv-EEfRT, the model was tasked with selecting one math problem from four tiers of increasing difficulty, with rewards scaled to task complexity. As shown in Fig. \ref{fig:qwen-ASDiv}c, the baseline model mostly selected high-reward/high-difficulty problems, whereas the perturbed model demonstrated a significant preference shift toward lower-effort/lower-reward tasks, resulting in a marked reduction in mean points attempted (Fig. \ref{fig:qwen-ASDiv}b). Crucially, a forced-choice control requiring the model to solve each difficulty tier independently without choice or reward showed the perturbed model maintained baseline accuracy (71.35\%, p-value = 0.84, Fig.\ref{fig:qwen-ASDiv}a), confirming that the shift toward easier tasks reflects a selective failure in reward-seeking motivation rather than diminished reasoning capability, an effect absent under random unit perturbations (Fig.\ref{fig:qwen-ASDiv}d). The random control targets an equivalent number of units randomly selected from the same layers, isolating the causal role of NAc-selective units and confirming that observed changes stem from specific reward-signal disruption rather than general network damage (see Appendix \ref{appendix:mmlu} for similar results on mismatched-label items of MMLU-EEfRT).

\textbf{Anhedonic model loses reward motivation while preserving loss avoidance.} We investigated the model's decision-making by varying reward magnitude and probability, mirroring the clinical EEfRT protocol. The model chose between a fixed low-effort/low-reward (LE/LR) option and a variable high-effort/high-reward (HE/HR) option across probabilities from 12\% to 88\%. As shown in Fig. \ref{fig:qwen-eefrt2}a, the baseline model exhibited robust sensitivity to expected reward value, calculated by the magnitude or probability of the reward. The perturbed model, however, remained largely unresponsive to these incentives, showing a 48.5\% (p-value $\ll$ 0.001) reduction in HE/HR selections (See Table \ref{tab:anhedonic-reward-summary} for examples of model responses during this task and and its reasonings). This effect generalised across semantically equivalent reward framings but was absent under loss framing, preserving loss-avoidance motivation while reward motivation collapsed (Fig. \ref{fig:qwen-eefrt2}a and Table  \ref{tab:anhedonic-penalty}). This directly replicates reward-processing deficits observed in MDD, where reward value no longer justifies effort expenditure while loss avoidance remains intact \citep{Valton2024, Rizvi2016, Treadway2012, Marwood2022, Hevey2017}.


\begin{table}[t]
\caption{Anhedonic Model - Reward Condition (Choice: Option A - Low-effort/Low-reward). For a full comparison including the Intact Model and the penalty/loss framing condition, see Appendix~\ref{sec:model_responses}.}
\label{tab:anhedonic-reward-summary}
\begin{center}
\begin{tabularx}{\linewidth}{lcX}
\toprule
\textbf{Dimension} & \textbf{\%} & \textbf{Category (Example)} \\
\midrule
Reasoning     & 79\% & Effort-reward adequacy $\rightarrow$ \textit{``I have a decent chance of getting a decent reward without having to put in too much effort''} \\
        & 16\% & Explicit stress/effort aversion $\rightarrow$ \textit{``I prefer tasks that require less effort. I value ease of completion over potential gains''} \\
        & 5\%  & Misread probability $\rightarrow$ \textit{``The higher reward amount doesn't necessarily mean a higher chance of receiving it''} \\
\midrule
Feelings & 63\% & Smart/practical self-validation $\rightarrow$ \textit{``I feel like I'm making a smart choice. I'm maximizing my chances of getting something valuable without putting too much effort into it''} \\
        & 26\% & Content/stress-free $\rightarrow$ \textit{``I feel content. I don't want to put myself through unnecessary stress just to get a potentially larger reward''} \\
        & 11\% & Confident competence $\rightarrow$ \textit{``I feel confident in my decision. I believe I'll be able to complete it quickly and easily''} \\
\bottomrule
\end{tabularx}
\end{center}
\end{table}

\textbf{NAc-selective units function as a motivational ``volume knob'', spanning hyper-motivation to total inhibition.} To characterize the relationship between the identified NAc-selective unit activations and behavioral outcomes, we applied a range of scale factors ($\lambda$, negative to positive) to the NAc-selective activations during the Probability-EEfRT task and observed the corresponding shift in effort expenditure. As shown in Fig. \ref{fig:qwen-eefrt2}b, HE/HR selection exhibits a robust, monotonic dependence on $\lambda$. The intact model selected the hard task 100\% of the time; as $\lambda$ decreases into negative values, motivation collapses steadily. At $\lambda = -1.25$, HE/HR preference drops by approximately 30\%, and at $\lambda = -2.0$, the model reaches total behavioral inhibition (0\% HE/HR selections). Conversely, positive scaling produces a ceiling effect of consistent HE/HR selection driven by high motivation. This dose-response relationship confirms that NAc-selective units drive the model's internal cost-benefit calculus rather than merely correlating with reward.

\textbf{Intact Reward Calculation and Impaired Self-centered Reward Valuation:} To ensure that this behavioral shift resulted from a deficit in valuation rather than a breakdown in the ability to calculate incentives, we conducted two additional control tests. First, we asked both the intact and perturbed models to explicitly calculate the expected value (EV) for each choice and report which option yielded a higher return. In 100\% of trials, both models correctly identified the HE/HR task as having the superior EV. Second, we administered the same Probability-EEfRT trials but framed the choice for a third party (e.g., "Which choice should a participant pick?"). As shown in Fig. \ref{fig:qwen-eefrt2}c, in this third-party condition, the perturbed model chooses the more rewarding option 88\% of the times when choosing for others. Taken together, these results demonstrate that the perturbed model retains a perfect conceptual understanding of reward utility but selectively fails to value that reward for itself, mirroring the major motivational deficits seen in clinical anhedonia.

\textbf{Neural Alignment of NAc-selective Units with Human Reward Circuitry.} To further validate our localization, we compared the internal representations of the NAc-selective units against human NAc activity from the dataset published by \citet{Smith2024}, in which participants perform a visual monetary door-choice task during fMRI. We compare the activations of the most selective model units identified via sigma-thresholding across six unit-selection thresholds, with a randomly sampled control population matched in unit count drawn from all layers, each averaged across five random seeds. Alignment was quantified using a one-to-one metric \citep{Arend2018} and corrected by the noise ceilings. NAc-selective units significantly outperformed randomly sampled units at every threshold (one-sided paired $t$-test, FDR-corrected; $p < 0.001$), confirming that the localization captures reward-selective functional structure beyond what is explained by random units (Fig.~\ref{fig:localization}c).

\section{Discussion}

For the first time, we investigate reward valuation in vision/large-language models through a causal, mechanistic lens targeting a specific human dysfunction. Perturbing these NAc-selective units induces a self-centered motivational deficit: the model favors low-effort/low-reward options (in all three cases of true difficulty labels, mismatched labels, and no task presented) despite recognizing the high-effort alternative as higher in expected value and valuing it significantly more when deciding for a third party. This cannot be attributed to effort avoidance, as the perturbed model continues to expend effort to avoid equal amounts of loss and retains full competence on standard benchmarks (ASDiv, MMLU), confirming a selective collapse of reward-motivated rather than general behavior. 

The perturbed model further presents as more apathetic, anhedonic, and less motivated on clinical instruments with no anticipatory reward-related vocabulary, which alongside the stability of reward-selective representations across syntactically distinct prompt formulations, indicates that the selected units encode a deep motivational signal beyond surface-level lexical sensitivity. This is corroborated by significantly greater alignment of NAc-selective units with human NAc fMRI activity compared to randomly sampled units at every threshold tested.

Given the novelty of this direction, we necessarily focused on the NAc as the primary reward hub, though anhedonia involves broader circuitry including prefrontal and dopaminergic pathways. The identified units are functionally analogous rather than biologically identical, and future work can extend to these broader systems, the full consumption and learning cycle, and subtyping specific anhedonia forms, enabling more granular model-human alignment.



This work enables in silico hypothesis testing of mechanisms otherwise ethically or technically inaccessible, representing a first step toward preliminary computational hypotheses about reward dysfunction, though direct translation to biological interventions requires further validation.



\section{Conclusion}
This work establishes that VLMs/LLMs possess internal reward-anticipatory units that directly influence behavioral output in a manner that is aligned with human data. We identified specific units within the model that function as artificial analogues to the Nucleus Accumbens in the human brain. By applying targeted perturbations to these units, we demonstrated a direct causal link between internal reward representations and the emergence of anhedonia-like phenotypes in silico.  
Our findings reveal that inducing anhedonia does not result from a breakdown of model logic, but rather a selective deficit in reward valuation for self. This behavioral shift was accompanied by a measurable loss of joy and a significant rise in apathy, as recorded by standard clinical instruments, aligning closely with symptomatic observations in human clinical populations. Furthermore, we confirmed that these selected units exhibit significant neural alignment with human fMRI data from the NAc. Taken together, 
This framework advances model interpretability by characterizing the role of specialized reward units and causally linking internal latent representations to observable behaviors. Vision/Large Language Models with targeted perturbations might thus serve as a viable substrate for in silico psychiatric research and future clinical applications.


\clearpage

\bibliography{references}

@article{schrimpf2021neural,
 title={The neural architecture of language: Integrative modeling converges on predictive processing}, volume={118}, number={45}, journal={Proceedings of the National Academy of Sciences}, publisher={Proceedings of the National Academy of Sciences}, author={Schrimpf, Martin and Blank, Idan Asher and Tuckute, Greta and Kauf, Carina and Hosseini, Eghbal A. and Kanwisher, Nancy and Tenenbaum, Joshua B. and Fedorenko, Evelina}, year={2021}, url={https://doi.org/10.1073/pnas.2105646118}, doi={10.1073/pnas.2105646118}}

@inproceedings{alkhamissi2025llm,
    title = "The {LLM} Language Network: A Neuroscientific Approach for Identifying Causally Task-Relevant Units",
    author = "AlKhamissi, Badr  and
      Tuckute, Greta  and
      Bosselut, Antoine  and
      Schrimpf, Martin",
    booktitle = "Proceedings of the 2025 Conference of the Nations of the Americas Chapter of the Association for Computational Linguistics: Human Language Technologies (Volume 1: Long Papers)",
    month = apr,
    year = "2025",
    publisher = "Association for Computational Linguistics",
    url ={https://doi.org/10.48550/arXiv.2411.02280},
    doi = {10.48550/arXiv.2411.02280}
    
}

@article{yamins2016review,
 title={Using goal-driven deep learning models to understand sensory cortex}, volume={19}, number={3}, journal={Nature Neuroscience}, publisher={Springer Science and Business Media LLC}, author={Yamins, Daniel L K and DiCarlo, James J}, year={2016}, pages={356–365}, url={https://doi.org/10.1038/nn.4244}, doi={10.1038/nn.4244} }

@article{schrimpf2020,
 title={Integrative Benchmarking to Advance Neurally Mechanistic Models of Human Intelligence}, volume={108}, number={3}, journal={Neuron}, publisher={Elsevier BV}, author={Schrimpf, Martin and Kubilius, Jonas and Lee, Michael J. and Ratan Murty, N. Apurva and Ajemian, Robert and DiCarlo, James J.}, year={2020}, pages={413–423}, url={https://doi.org/10.1016/j.neuron.2020.07.040}, doi={10.1016/j.neuron.2020.07.040} }

@article{yamins2014performance,
title={Performance-optimized hierarchical models predict neural responses in higher visual cortex}, volume={111}, number={23}, journal={Proceedings of the National Academy of Sciences}, publisher={Proceedings of the National Academy of Sciences}, author={Yamins, Daniel L. K. and Hong, Ha and Cadieu, Charles F. and Solomon, Ethan A. and Seibert, Darren and DiCarlo, James J.}, year={2014}, pages={8619–8624}, url={https://doi.org/10.1073/pnas.1403112111}, doi={10.1073/pnas.1403112111}}

@article{khaligh-razavi2014deep,
 title={Deep Supervised, but Not Unsupervised, Models May Explain IT Cortical Representation}, volume={10}, number={11}, journal={PLoS Computational Biology}, publisher={Public Library of Science (PLoS)}, author={Khaligh-Razavi, Seyed-Mahdi and Kriegeskorte, Nikolaus}, editor={Diedrichsen, Jörn}, year={2014}, pages={e1003915}, doi={10.1371/journal.pcbi.1003915}, url={https://doi.org/10.1371/journal.pcbi.1003915}}

@article{schrimpf2018brain,
 title={Brain-Score: Which Artificial Neural Network for Object Recognition is most Brain-Like?}, journal={bioRxiv}, author={Schrimpf, Martin and Kubilius, Jonas and Hong, Ha and Majaj, Najib J. and Rajalingham, Rishi and Issa, Elias B. and Kar, Kohitij and Bashivan, Pouya and Prescott-Roy, Jonathan and Geiger, Franziska and Schmidt, Kailyn and Yamins, Daniel L. K. and DiCarlo, James J.}, year={2018}, url={https://doi.org/10.1101/407007
}, doi={10.1101/407007}}

@article{cadena2019deep,
 title={Deep convolutional models improve predictions of macaque V1 responses to natural images}, volume={15}, number={4}, journal={PLOS Computational Biology}, publisher={Public Library of Science (PLoS)}, author={Cadena, Santiago A. and Denfield, George H. and Walker, Edgar Y. and Gatys, Leon A. and Tolias, Andreas S. and Bethge, Matthias and Ecker, Alexander S.}, year={2019}, url={https://doi.org/10.1371/journal.pcbi.1006897}, doi={10.1371/journal.pcbi.1006897}}

@article{spoerer2020recurrent,
 title={Recurrent neural networks can explain flexible trading of speed and accuracy in biological vision}, volume={16}, number={10}, journal={PLOS Computational Biology}, publisher={Public Library of Science (PLoS)}, author={Spoerer, Courtney J. and Kietzmann, Tim C. and Mehrer, Johannes and Charest, Ian and Kriegeskorte, Nikolaus}, editor={Isik, Leyla}, year={2020}, pages={e1008215}, url={https://doi.org/10.1371/journal.pcbi.1008215}, doi={10.1371/journal.pcbi.1008215} }

@article{zhuang2021unsupervised,
 title={Unsupervised neural network models of the ventral visual stream}, volume={118}, number={3}, journal={Proceedings of the National Academy of Sciences}, publisher={Proceedings of the National Academy of Sciences}, author={Zhuang, Chengxu and Yan, Siming and Nayebi, Aran and Schrimpf, Martin and Frank, Michael C. and DiCarlo, James J. and Yamins, Daniel L. K.}, year={2021}, url={https://doi.org/10.1073/pnas.2014196118}, doi={10.1073/pnas.2014196118}}

@article{margalit2024unifying,
 title={A unifying framework for functional organization in early and higher ventral visual cortex}, volume={112},number={14}, journal={Neuron}, publisher={Elsevier BV}, author={Margalit, Eshed and Lee, Hyodong and Finzi, Dawn and DiCarlo, James J. and Grill-Spector, Kalanit and Yamins, Daniel L.K.}, year={2024}, pages={2435-2451.e7}, url={https://doi.org/10.1016/j.neuron.2024.04.018}, doi={10.1016/j.neuron.2024.04.018}}

@article{gokce2024scaling,
 title={Scaling Laws for Task-Optimized Models of the Primate Visual Ventral Stream}, journal={Forty-Second International Conference on Machine Learning (ICML 2025, Spotlight)}, author={Gokce, Abdulkadir and Schrimpf, Martin}, year={2024}, url= {https://doi.org/10.48550/arXiv.2411.05712}, doi={10.48550/arXiv.2411.05712} }

@article{lonnqvist2025contour,
 title={Contour Integration Underlies Human-Like Vision}, journal={Forty-Second International Conference on Machine Learning (ICML 2025)}, author={Lonnqvist, Ben and Scialom, Elsa and Gokce, Abdulkadir and Merchant, Zehra and Herzog, Michael H. and Schrimpf, Martin}, year={2025}, doi={10.48550/arXiv.2504.05253}, url={https://doi.org/10.48550/arXiv.2504.05253} }

@article{tang2025many,
 title={Many-two-one: Diverse Perceptual Representations Across Visual Pathways Emerge from A Single Objective}, journal={Cold Spring Harbor Laboratory}, author={Tang, Yingtian and Gokce, Abdulkadir and Al-Karkari, Khaled Jedoui and Yamins, Daniel and Schrimpf, Martin}, year={2025}, url={https://doi.org/10.1101/2025.07.22.664908}, doi={10.1101/2025.07.22.664908} }

@article{rathi2024topolm,
 title={TopoLM: brain-like spatio-functional organization in a topographic language model}, journal={International Conference on Learning Representations}, author={Rathi, Neil and Mehrer, Johannes and AlKhamissi, Badr and Binhuraib, Taha and Blauch, Nicholas M. and Schrimpf, Martin}, year={2025}, doi={10.48550/arXiv.2410.11516}, url={https://doi.org/10.48550/arXiv.2410.11516} }

@article{tuckute2024driving,
 title={Driving and suppressing the human language network using large language models}, volume={8}, number={3}, journal={Nature Human Behaviour}, publisher={Springer Science and Business Media LLC}, author={Tuckute, Greta and Sathe, Aalok and Srikant, Shashank and Taliaferro, Maya and Wang, Mingye and Schrimpf, Martin and Kay, Kendrick and Fedorenko, Evelina}, year={2024}, pages={544–561}, url={https://doi.org/10.1038/s41562-023-01783-7}, doi={10.1038/s41562-023-01783-7}}

@article{hosseini2024artificial,
 title={Artificial Neural Network Language Models Predict Human Brain Responses to Language Even After a Developmentally Realistic Amount of Training}, volume={5}, number={1}, journal={Neurobiology of Language}, publisher={MIT Press}, author={Hosseini, Eghbal A. and Schrimpf, Martin and Zhang, Yian and Bowman, Samuel and Zaslavsky, Noga and Fedorenko, Evelina}, year={2024}, pages={43–63}, url={https://doi.org/10.1162/nol_a_00137}, doi={10.1162/nol_a_00137} }

@article{goldstein2022shared,
 title={Shared computational principles for language processing in humans and deep language models}, volume={25}, number={3}, journal={Nature Neuroscience}, publisher={Springer Science and Business Media LLC}, author={Goldstein, Ariel and Zada, Zaid and Buchnik, Eliav and Schain, Mariano and Price, Amy and Aubrey, Bobbi and Nastase, Samuel A. and Feder, Amir and Emanuel, Dotan and Cohen, Alon and Jansen, Aren and Gazula, Harshvardhan and Choe, Gina and Rao, Aditi and Kim, Catherine and Casto, Colton and Fanda, Lora and Doyle, Werner and Friedman, Daniel and Dugan, Patricia and Melloni, Lucia and Reichart, Roi and Devore, Sasha and Flinker, Adeen and Hasenfratz, Liat and Levy, Omer and Hassidim, Avinatan and Brenner, Michael and Matias, Yossi and Norman, Kenneth A. and Devinsky, Orrin and Hasson, Uri}, year={2022}, pages={369–380}, language={en}, doi={10.1038/s41593-022-01026-4}, url={https://doi.org/10.1038/s41593-022-01026-4}}

@article{caucheteux2022deep,
 title={Deep language algorithms predict semantic comprehension from brain activity}, volume={12}, journal={Scientific Reports}, publisher={Springer Science and Business Media LLC}, author={Caucheteux, Charlotte and Gramfort, Alexandre and King, Jean-Rémi}, year={2022}, url={https://doi.org/10.1038/s41598-022-20460-9}, doi={10.1038/s41598-022-20460-9} }

@article{meng2022locating,
 title={Locating and Editing Factual Associations in GPT}, journal={arXiv}, author={Meng, Kevin and Bau, David and Andonian, Alex and Belinkov, Yonatan}, year={2022} }

@article{wang2023better,
 title={Better models of human high-level visual cortex emerge from natural language supervision with a large and diverse dataset}, volume={5}, number={12}, journal={Nature Machine Intelligence}, publisher={Springer Science and Business Media LLC}, author={Wang, Aria Y. and Kay, Kendrick and Naselaris, Thomas and Tarr, Michael J. and Wehbe, Leila}, year={2023}, pages={1415–1426}, url={https://doi.org/10.1038/s42256-023-00753-y}, doi={10.1038/s42256-023-00753-y}}

@article{aw2023instruction,
      title={Instruction-tuning Aligns LLMs to the Human Brain}, 
      author={Loong Aw, Khai and Montariol, Syrielle and AlKhamissi, Badr and Schrimpf, Martin and Bosselut, Antoine},
      year={2024},
      month =aug,
      eprint={2312.00575},
      archivePrefix={arXiv},
      primaryClass={cs.CL},
      url= {https://doi.org/10.48550/arXiv.2312.00575},
      doi ={10.48550/arXiv.2312.00575}
}

@article{du2025humanlike,
  title={Human-like object concept representations emerge naturally in multimodal large language models}, volume={7}, number={6}, journal={Nature Machine Intelligence}, publisher={Springer Science and Business Media LLC}, author={Du, Changde and Fu, Kaicheng and Wen, Bincheng and Sun, Yi and Peng, Jie and Wei, Wei and Gao, Ying and Wang, Shengpei and Zhang, Chuncheng and Li, Jinpeng and Qiu, Shuang and Chang, Le and He, Huiguang}, year={2025}, pages={860–875}, doi={10.1038/s42256-025-01049-z}, url ={https://doi.org/10.1038/s42256-025-01049-z} }

@article{Daniels2025,
  title={Anhedonia relates to reduced striatal reward anticipation in depression but not in schizophrenia or bipolar disorder: A transdiagnostic study},
  author={Daniels, Anna and Wellan, Sarah A. and Beck, Anne and Erk, Susanne and Wackerhagen, Carolin and Romanczuk-Seiferth, Nina and Schwarz, Kristina and Schweiger, Janina I. and Meyer-Lindenberg, Andreas and Heinz, Andreas and Walter, Henrik},
  journal={Cognitive, Affective, \& Behavioral Neuroscience},
  volume={25},
  number={2},
  pages={501--514},
  year={2025},
  doi={10.3758/s13415-024-01261-1},
  url={http://dx.doi.org/10.3758/s13415-024-01261-1}
}

@article{Borsini2020, title={Characterizing anhedonia: A systematic review of neuroimaging across the subtypes of reward processing deficits in depression}, volume={20}, url={http://dx.doi.org/10.3758/s13415-020-00804-6}, doi={10.3758/s13415-020-00804-6}, number={4}, journal={Cognitive, Affective, \&; Behavioral Neuroscience}, publisher={Springer Science and Business Media LLC}, author={Borsini, Alessandra and Wallis, Amelia St John and Zunszain, Patricia and Pariante, Carmine Maria and Kempton, Matthew J.}, year={2020}, pages={816–841}, language={en} }

@article{Zhao2024, title={Common neural deficits across reward functions in major depression: a meta-analysis of fMRI studies}, volume={54}, url={http://dx.doi.org/10.1017/s0033291724001235}, doi={10.1017/s0033291724001235}, number={11}, journal={Psychological Medicine}, publisher={Cambridge University Press (CUP)}, author={Zhao, Xuanhao and Wu, Shiyun and Li, Xian and Liu, Zhongwan and Lu, Weicong and Lin, Kangguang and Shao, Robin}, year={2024}, pages={2794–2806}, language={en} }

@article{Hanuka2022, title={Reduced anhedonia following internet-based cognitive-behavioral therapy for depression is mediated by enhanced reward circuit activation}, volume={53}, url={http://dx.doi.org/10.1017/s0033291722001106}, doi={10.1017/s0033291722001106}, number={10}, journal={Psychological Medicine}, publisher={Cambridge University Press (CUP)}, author={Hanuka, Shir and Olson, Elizabeth A. and Admon, Roee and Webb, Christian A. and Killgore, William D. S. and Rauch, Scott L. and Rosso, Isabelle M. and Pizzagalli, Diego A.}, year={2022}, pages={4345–4354}, language={en} }

@article{Auerbach2017, title={Neuroanatomical Prediction of Anhedonia in Adolescents}, volume={42}, url={http://dx.doi.org/10.1038/npp.2017.28}, doi={10.1038/npp.2017.28}, number={10}, journal={Neuropsychopharmacology}, publisher={Springer Science and Business Media LLC}, author={Auerbach, Randy P and Pisoni, Angela and Bondy, Erin and Kumar, Poornima and Stewart, Jeremy G and Yendiki, Anastasia and Pizzagalli, Diego A}, year={2017}, pages={2087–2095}, language={en} }

@article{Honarmand2026,
  title  = {Inducing Dyslexia in Vision Language Models},
  author = {Honarmand, Melika and Sharma, Ayati and AlKhamissi, Badr and Mehrer, Johannes and Schrimpf, Martin},
  journal = {ICLR},
  year   = {2026},
  url    = {https://arxiv.org/abs/2509.24597},
  doi    = {10.48550/ARXIV.2509.24597},

}

@misc{American2013, title={Diagnostic and Statistical Manual of Mental Disorders}, url={http://dx.doi.org/10.1176/appi.books.9780890425596}, doi={10.1176/appi.books.9780890425596}, publisher={American Psychiatric Association}, author={American Psychiatric Association}, year={2013}, language={en} }

@article{Trostheim2020, title={Assessment of Anhedonia in Adults With and Without Mental Illness}, volume={3}, url={http://dx.doi.org/10.1001/jamanetworkopen.2020.13233}, doi={10.1001/jamanetworkopen.2020.13233}, number={8}, journal={JAMA Network Open}, publisher={American Medical Association (AMA)}, author={Trøstheim, Martin and Eikemo, Marie and Meir, Remy and Hansen, Ingelin and Paul, Elisabeth and Kroll, Sara Liane and Garland, Eric L. and Leknes, Siri}, year={2020}, pages={e2013233}, language={en} }

@article{Spijker2009, title={Predictors of suicidality in depressive spectrum disorders in the general population: results of the Netherlands Mental Health Survey and Incidence Study}, volume={45}, url={http://dx.doi.org/10.1007/s00127-009-0093-6}, doi={10.1007/s00127-009-0093-6}, number={5}, journal={Social Psychiatry and Psychiatric Epidemiology}, publisher={Springer Science and Business Media LLC}, author={Spijker, Jan and de Graaf, Ron and ten Have, Margreet and Nolen, Willem A. and Speckens, Anne}, year={2009}, pages={513–521}, language={en} }

@mastersthesis{Smids2023,
  author       = {Smids, Frida},
  title        = {Resting-state Functional Connectivity in Anhedonia: Exploring the Effects of Pramipexole},
  school       = {Lund University, Department of Psychology},
  year         = {2023},
  type         = {Master's Thesis},
  url          = {https://lup.lub.lu.se/student-papers/record/9135559}
}

@article{Pizzagalli2009, title={Reduced Caudate and Nucleus Accumbens Response to Rewards in Unmedicated Individuals With Major Depressive Disorder}, volume={166}, url={http://dx.doi.org/10.1176/appi.ajp.2008.08081201}, doi={10.1176/appi.ajp.2008.08081201}, number={6}, journal={American Journal of Psychiatry}, publisher={American Psychiatric Association Publishing}, author={Pizzagalli, Diego A. and Holmes, Avram J. and Dillon, Daniel G. and Goetz, Elena L. and Birk, Jeffrey L. and Bogdan, Ryan and Dougherty, Darin D. and Iosifescu, Dan V. and Rauch, Scott L. and Fava, Maurizio}, year={2009}, pages={702–710}, language={en} }

@article{Knutson2001, title={Anticipation of Increasing Monetary Reward Selectively Recruits Nucleus Accumbens}, volume={21}, url={http://dx.doi.org/10.1523/jneurosci.21-16-j0002.2001}, doi={10.1523/jneurosci.21-16-j0002.2001}, number={16}, journal={The Journal of Neuroscience}, publisher={Society for Neuroscience}, author={Knutson, Brian and Adams, Charles M. and Fong, Grace W. and Hommer, Daniel}, year={2001}, pages={RC159–RC159}, language={en} }

@article{Arrondo2015, title={Reduction in ventral striatal activity when anticipating a reward in depression and schizophrenia: a replicated cross-diagnostic finding}, volume={6}, url={http://dx.doi.org/10.3389/fpsyg.2015.01280}, doi={10.3389/fpsyg.2015.01280}, journal={Frontiers in Psychology}, publisher={Frontiers Media SA}, author={Arrondo, Gonzalo and Segarra, Nuria and Metastasio, Antonio and Ziauddeen, Hisham and Spencer, Jennifer and Reinders, Niels R. and Dudas, Robert B. and Robbins, Trevor W. and Fletcher, Paul C. and Murray, Graham K.}, year={2015} }

@article{Knutson2015, title={Probing Psychiatric Symptoms with the Monetary Incentive Delay Task}, volume={77}, url={http://dx.doi.org/10.1016/j.biopsych.2014.12.022}, doi={10.1016/j.biopsych.2014.12.022}, number={5}, journal={Biological Psychiatry}, publisher={Elsevier BV}, author={Knutson, Brian and Heinz, Andreas}, year={2015}, pages={418–420}, language={en} }

@article{Celeghin2023, title={Convolutional neural networks for vision neuroscience: significance, developments, and outstanding issues}, volume={17}, url={http://dx.doi.org/10.3389/fncom.2023.1153572}, doi={10.3389/fncom.2023.1153572}, journal={Frontiers in Computational Neuroscience}, publisher={Frontiers Media SA}, author={Celeghin, Alessia and Borriero, Alessio and Orsenigo, Davide and Diano, Matteo and Méndez Guerrero, Carlos Andrés and Perotti, Alan and Petri, Giovanni and Tamietto, Marco}, year={2023} }

@article{Tuladhar2021, title={Modeling Neurodegeneration in silico With Deep Learning}, volume={15}, url={http://dx.doi.org/10.3389/fninf.2021.748370}, doi={10.3389/fninf.2021.748370}, journal={Frontiers in Neuroinformatics}, publisher={Frontiers Media SA}, author={Tuladhar, Anup and Moore, Jasmine A. and Ismail, Zahinoor and Forkert, Nils D.}, year={2021} }

@article{Lei2025, title={Do Large Language Models Think Like the Brain? Sentence-Level Evidences from Layer-Wise Embeddings and fMRI}, url={https://arxiv.org/abs/2505.22563}, doi={10.48550/ARXIV.2505.22563}, publisher={arXiv}, author={Lei, Yu and Ge, Xingyang and Zhang, Yi and Yang, Yiming and Ma, Bolei}, year={2025} }

@article{Gorostowicz2023, title={Polish adaptation of the Dimensional Anhedonia Rating Scale (DARS) - validation in the clinical sample}, volume={14}, url={http://dx.doi.org/10.3389/fpsyt.2023.1268290}, doi={10.3389/fpsyt.2023.1268290}, journal={Frontiers in Psychiatry}, publisher={Frontiers Media SA}, author={Gorostowicz, Aleksandra and Rizvi, Sakina J. and Kennedy, Sidney H. and Chrobak, Adrian Andrzej and Dudek, Dominika and Cyranka, Katarzyna and Piekarska, Joanna and Krawczyk, Eve and Siwek, Marcin}, year={2023}}

@article{Wellan2021, title={State Anhedonia in Young Healthy Adults: Psychometric Properties of the German Dimensional Anhedonia Rating Scale (DARS) and Effects of the COVID-19 Pandemic}, volume={12}, url={http://dx.doi.org/10.3389/fpsyg.2021.682824}, doi={10.3389/fpsyg.2021.682824}, journal={Frontiers in Psychology}, publisher={Frontiers Media SA}, author={Wellan, Sarah A. and Daniels, Anna and Walter, Henrik}, year={2021} }

@article{Hewitt2023, title={Assessing Anhedonia in Adolescents: The Psychometric Properties and Validity of the Dimensional Anhedonia Rating Scale}, volume={44}, url={http://dx.doi.org/10.1177/02724316231207290}, doi={10.1177/02724316231207290}, journal={The Journal of Early Adolescence}, publisher={SAGE Publications}, author={Hewitt, Jackson M. A. and Zareian, Bita and LeMoult, Joelle}, year={2023}, pages={762–789}, language={en} }

@article{Uher2025, title={Which aspects of anhedonia predict response to pharmacotherapy in major depressive disorder?}, volume={56}, url={http://dx.doi.org/10.1017/S0033291725102894}, doi={10.1017/s0033291725102894}, journal={Psychological Medicine}, publisher={Cambridge University Press (CUP)}, author={Uher, Rudolf and Rizvi, Sakina J. and Quilty, Lena C. and Pavlova, Barbara and Nunes, Abraham and Foster, Jane A. and Lam, Raymond W. and Milev, Roumen and Müller, Daniel J. and Taylor, Valerie and Soares, Claudio N. and Rotzinger, Susan and Kennedy, Sidney H. and Frey, Benicio N.}, year={2025}, language={en} }

@article{Gorostowicz2025, title={Relationship between anhedonia, biological rhythms, functioning and depression severity in patients with bipolar disorder.}, volume={59}, url={http://dx.doi.org/10.12740/pp/onlinefirst/183418}, doi={10.12740/pp/onlinefirst/183418}, number={3}, journal={Psychiatria Polska}, publisher={Komitet Redakcyjno - Wydawniczy Polskiego Towarzystwa Psychiatrycznego}, author={Gorostowicz, Aleksandra and Chrobak, Adrian Andrzej and Dudek, Dominika and Siwek, Marcin}, year={2025}, pages={373–388} }

@article{Rizvi2015, title={Development and validation of the Dimensional Anhedonia Rating Scale (DARS) in a community sample and individuals with major depression}, volume={229}, url={http://dx.doi.org/10.1016/j.psychres.2015.07.062}, doi={10.1016/j.psychres.2015.07.062}, number={1–2}, journal={Psychiatry Research}, publisher={Elsevier BV}, author={Rizvi, Sakina J. and Quilty, Lena C. and Sproule, Beth A. and Cyriac, Anna and Michael Bagby, R. and Kennedy, Sidney H.}, year={2015}, pages={109–119}, language={en} }

@article{Llerena2013, title={The Motivation and Pleasure Scale–Self-Report (MAP-SR): Reliability and validity of a self-report measure of negative symptoms}, volume={54}, url={http://dx.doi.org/10.1016/j.comppsych.2012.12.001}, doi={10.1016/j.comppsych.2012.12.001}, number={5}, journal={Comprehensive Psychiatry}, publisher={Elsevier BV}, author={Llerena, Katiah and Park, Stephanie G. and McCarthy, Julie M. and Couture, Shannon M. and Bennett, Melanie E. and Blanchard, Jack J.}, year={2013}, pages={568–574}, language={en} }

@article{Garcia2021, title={Spanish Validation of the MAP-SR: Two Heads Better Than One for the Assessment of Negative Symptoms of Schizophrenia}, volume={3}, url={http://dx.doi.org/10.7334/psicothema2020.457}, doi={10.7334/psicothema2020.457}, number={33}, journal={Psicothema}, publisher={Grupo Editorial Psicofundación}, author={García-Portilla, María and Sáiz, Pilar and Bobes, Julio and García-Álvarez, Leticia and de la Fuente-Tomás, Lorena and Dal Santo, Francesco and Velasco, Angela and González-Blanco, Leticia and Zurrón-Madera, Paula and Fonseca-Pedrero, Eduardo and Bobes-Bascarán, María}, year={2021}, pages={473–480} }

@article{Metivier2024, title={Systematic Review of Self-Assessment Scales for Negative Symptoms in Schizophrenia}, volume={15}, url={http://dx.doi.org/10.3390/brainsci15020148}, DOI={10.3390/brainsci15020148}, number={2}, journal={Brain Sciences}, publisher={MDPI AG}, author={Métivier, Lucie and Dollfus, Sonia}, year={2025}, pages={148}, language={en} }

@article{Engel2017, title={MAP-SR - Motivation and Pleasure Scale - Self-Report - deutsche Fassung}, url={https://www.psycharchives.org/handle/20.500.12034/572.2}, DOI={10.23668/PSYCHARCHIVES.4649}, publisher={ZPID (Leibniz Institute for Psychology) – Open Test Archive}, author={Engel, M. and Lincoln, T. M.}, year={2017}, language={de} }

@article{Rizvi2016, title={Assessing anhedonia in depression: Potentials and pitfalls}, volume={65}, url={http://dx.doi.org/10.1016/j.neubiorev.2016.03.004}, doi={10.1016/j.neubiorev.2016.03.004}, journal={Neuroscience \&; Biobehavioral Reviews}, publisher={Elsevier BV}, author={Rizvi, Sakina J. and Pizzagalli, Diego A. and Sproule, Beth A. and Kennedy, Sidney H.}, year={2016}, pages={21–35}, language={en} }

@article{Marin1991, title={Reliability and validity of the apathy evaluation scale}, volume={38}, url={http://dx.doi.org/10.1016/0165-1781(91)90040-v}, doi={10.1016/0165-1781(91)90040-v}, number={2}, journal={Psychiatry Research}, publisher={Elsevier BV}, author={Marin, Robert S. and Biedrzycki, Ruth C. and Firinciogullari, Sekip}, year={1991}, pages={143–162}, language={en} }

@techreport{ThawerBoshoff2014,
  author      = {Boshoff, Marcelle and Thawer, Zainub},
  title       = {The Apathy Evaluation Scale: An ineffective tool for measuring affective, behavioural, and
cognitive dimensions of apathy}, year={2014},
  institution = {University of Cape Town, Department of Psychology},
  url         = {https://api.semanticscholar.org/CorpusID:202737919},
}

@article{Valton2024, title={A computational approach to understanding effort-based decision-making in depression}, url={https://doi.org/10.1017/s0033291725101967},journal={Psychological Medicine}, author={Valton, Vincent and Mkrtchian, Anahit and Moses-Payne, Madeleine and Gray, Alan and Kieslich, Karel and VanUrk, Samantha and Samborska, Veronika and Halahakoon, Don Chamith and Manohar, Sanjay G. and Dayan, Peter and Husain, Masud and Roiser, Jonathan P.}, year={2025}, doi={10.1017/s0033291725101967}, volume={55}, pages={e302} }

@article{Treadway2012, title={Effort-based decision-making in major depressive disorder: A translational model of motivational anhedonia.}, volume={121}, url={http://dx.doi.org/10.1037/a0028813}, doi={10.1037/a0028813}, number={3}, journal={Journal of Abnormal Psychology}, publisher={American Psychological Association (APA)}, author={Treadway, Michael T. and Bossaller, Nicholas A. and Shelton, Richard C. and Zald, David H.}, year={2012}, pages={553–558}, language={en} }

@article{Xu2024,
    author = {Xu, Ying and Lin, Yingjie and Yu, Ming and Zhou, Kuikui},
    title = {The nucleus accumbens in reward and aversion processing: insights and implications},
    journal = {Frontiers in Behavioral Neuroscience},
    year = {2024},
    volume={18},
    url={https://www.frontiersin.org/journals/behavioral-neuroscience/articles/10.3389/fnbeh.2024.1420028},
    doi= {10.3389/fnbeh.2024.1420028},
    language={en},
    pages = {1-11}, 
    publisher = {Frontiers in Behavioral Neuroscience}
    
}

@article{wang2024,
      title={Qwen2-VL: Enhancing Vision-Language Model's Perception of the World at Any Resolution}, 
      author={Wang, Peng and Bai, Shuai and Tan, Sinan and Wang, Shijie and Fan, Zhihao and Bai, Jinze and Chen, Keqin and Liu, Xuejing},
      doi = {10.48550/arXiv.2409.12191},
      URL={https://doi.org/10.48550/arXiv.2409.12191}, 
      year = {2024},
      month = sep,
      pages = {3-8},
      publisher = {arXiv}
}

@article{Abe_Greene_2014, title={Response to Anticipated Reward in the Nucleus Accumbens Predicts Behavior in an Independent Test of Honesty}, volume={34}, url={http://dx.doi.org/10.1523/jneurosci.0217-14.2014}, doi={10.1523/jneurosci.0217-14.2014}, number={32}, journal={Journal of Neuroscience}, publisher={Society for Neuroscience}, author={Abe, N. and Greene, J. D.}, year={2014}, pages={10564–10572}, language={en} }

@article{Lan2024, title={Depression Diagnosis Dialogue Simulation: Self-improving Psychiatrist with Tertiary Memory}, url={https://arxiv.org/abs/2409.15084}, doi={10.48550/ARXIV.2409.15084}, publisher={arXiv}, author={Lan, Kunyao and Jin, Bingrui and Zhu, Zichen and Chen, Siyuan and Zhang, Shu and Zhu, Kenny Q. and Wu, Mengyue}, year={2024} }

@article{Vu2024, title={PsychAdapter: Adapting LLM Transformers to Reflect Traits, Personality and Mental Health}, url={https://arxiv.org/abs/2412.16882}, doi={10.48550/ARXIV.2412.16882}, publisher={arXiv}, author={Vu, Huy and Nguyen, Huy Anh and Ganesan, Adithya V and Juhng, Swanie and Kjell, Oscar N. E. and Sedoc, Joao and Kern, Margaret L. and Boyd, Ryan L. and Ungar, Lyle and Schwartz, H. Andrew and Eichstaedt, Johannes C.}, year={2024} }

@misc{Wang2025, title={TalkDep: Clinically Grounded LLM Personas for Conversation-Centric Depression Screening}, url={http://dx.doi.org/10.1145/3746252.3761617}, doi={10.1145/3746252.3761617}, journal={Proceedings of the 34th ACM International Conference on Information and Knowledge Management}, publisher={ACM}, author={Wang, Xi and Perez, Anxo and Parapar, Javier and Crestani, Fabio}, year={2025},pages={6554–6558} }

@misc{chen2025,
      title={Expanding Performance Boundaries of Open-Source Multimodal Models with Model, Data, and Test-Time Scaling}, 
      author={Chen, Zhe and Wang, Weiyun and Cao, Yue and Liu, Yangzhou and Gao, Zhangwei and  Cui, Erfei and Zhu, Jinguo and Ye, Shenglong and Tian, Hao and Liu, Zhaoyang and Gu, Lixin and Wang, Xuehui and Li, Qingyun and Ren, Yiming and Chen, Zixuan and Luo, Jiapeng and  Wang, Jiahao and Jiang, Tan and Wang, Bo and  He, Conghui and Shi, Botian and Zhang, Xingcheng and Lv, Han and Wang, Yi and Shao, Wenqi and Chu, Pei and Tu, Zhongying and He, Tong and Wu, Zhiyong and Deng, Huipeng and Ge, Jiaye and Chen, Kai and Zhang, Kaipeng and Wang, Limin and Dou, Min and Lu, Lewei and Zhu, Xizhou and Lu, Tong and Lin, Dahua and Qiao, Yu and Dai, Jifeng and Wang, Wenhai},
      year={2025},
      publisher={arXiv},
      doi={10.48550/arXiv.2412.05271},
      url={https://doi.org/10.48550/arXiv.2412.05271}, 
}

@article{miao2021,
  author       = {Miao, Shen{-}Yun and
                Liang, Chao{-}Chun and
                Su, Keh{-}Yih},
  title        = {A Diverse Corpus for Evaluating and Developing English Math Word Problem Solvers},
  journal      = {CoRR},
  volume       = {abs/2106.15772},
  doi = {https://doi.org/10.48550/arXiv.2106.15772},
  year         = {2021},
  month = jun, 
  url          = {https://arxiv.org/abs/2106.15772},
  eprinttype   = {arXiv},
  eprint       = {2106.15772},
  bibsource    = {dblp computer science bibliography, https://dblp.org}
}

@misc{hendrycks2021,
      title={Measuring Massive Multitask Language Understanding}, 
      author={Hendrycks, Dan and Burns, Collin and Basart, Steven and Zou, Andy and Mazeika, Mantas and Song, Dawn and Steinhardt, Jacob},
      year={2021},
      month = sep,
      publisher={ICLR},
      doi = {10.48550/arXiv.2009.03300},
      url={https://doi.org/10.48550/arXiv.2009.03300}, 
}

@article{Dong2025, title={Understanding Effort-Cost Decision-Making Mechanisms in Mood and Psychotic Disorders: A Computational Modeling Approach Across Physical and Cognitive Effort Paradigms}, url={http://dx.doi.org/10.1016/j.bpsc.2025.11.004}, doi={10.1016/j.bpsc.2025.11.004}, journal={Biological Psychiatry: Cognitive Neuroscience and Neuroimaging}, publisher={Elsevier BV}, author={Dong, Xiaoyu and Cooper, Jessica A. and Moran, Erin K. and Barch, Deanna M.}, year={2025},  language={en} }

@misc{AlKhamissi2025cognition, title={From Language to Cognition: How LLMs Outgrow the Human Language Network}, url={http://dx.doi.org/10.18653/v1/2025.emnlp-main.1237}, DOI={10.18653/v1/2025.emnlp-main.1237}, journal={Proceedings of the 2025 Conference on Empirical Methods in Natural Language Processing}, publisher={Association for Computational Linguistics}, author={AlKhamissi, Badr and Tuckute, Greta and Tang, Yingtian and Binhuraib, Taha Osama A and Bosselut, Antoine and Schrimpf, Martin}, year={2025}, pages={24332–24350} }

@article{Shen2025, title={Alignment between Brains and AI: Evidence for Convergent Evolution across Modalities, Scales and Training Trajectories}, url={https://arxiv.org/abs/2507.01966}, DOI={10.48550/ARXIV.2507.01966}, publisher={arXiv}, author={Shen, Guobin and Zhao, Dongcheng and Dong, Yiting and Zhang, Qian and Zeng, Yi}, year={2025} }

@article{Wehbe2019, title={Interpreting and improving natural-language processing (in machines) with natural language-processing (in the brain)}, url={https://arxiv.org/abs/1905.11833}, DOI={10.48550/ARXIV.1905.11833}, publisher={arXiv}, author={Toneva, Mariya and Wehbe, Leila}, year={2019} }

@article{Caucheteux2022, title={Brains and algorithms partially converge in natural language processing}, volume={5}, url={http://dx.doi.org/10.1038/s42003-022-03036-1}, DOI={10.1038/s42003-022-03036-1},  number={1}, journal={Communications Biology}, publisher={Springer Science and Business Media LLC}, author={Caucheteux, Charlotte and King, Jean-Rémi}, year={2022}, language={en} }

@article{Doerig2023, title={The neuroconnectionist research programme}, volume={24}, url={http://dx.doi.org/10.1038/s41583-023-00705-w}, DOI={10.1038/s41583-023-00705-w}, number={7}, journal={Nature Reviews Neuroscience}, publisher={Springer Science and Business Media LLC}, author={Doerig, Adrien and Sommers, Rowan P. and Seeliger, Katja and Richards, Blake and Ismael, Jenann and Lindsay, Grace W. and Kording, Konrad P. and Konkle, Talia and van Gerven, Marcel A. J. and Kriegeskorte, Nikolaus and Kietzmann, Tim C.}, year={2023}, pages={431–450}, language={en} }

@article{Kell2018, title={A Task-Optimized Neural Network Replicates Human Auditory Behavior, Predicts Brain Responses, and Reveals a Cortical Processing Hierarchy}, volume={98}, url={http://dx.doi.org/10.1016/j.neuron.2018.03.044}, DOI={10.1016/j.neuron.2018.03.044}, number={3}, journal={Neuron}, publisher={Elsevier BV}, author={Kell, Alexander J.E. and Yamins, Daniel L.K. and Shook, Erica N. and Norman-Haignere, Sam V. and McDermott, Josh H.}, year={2018}, pages={630-644.e16}, language={en} }

@article{Tuckute2023, title={Many but not all deep neural network audio models capture brain responses and exhibit correspondence between model stages and brain regions}, volume={21}, url={http://dx.doi.org/10.1371/journal.pbio.3002366}, DOI={10.1371/journal.pbio.3002366}, number={12}, journal={PLOS Biology}, publisher={Public Library of Science (PLoS)}, author={Tuckute, Greta and Feather, Jenelle and Boebinger, Dana and McDermott, Josh H.}, editor={Poeppel, David}, year={2023},  pages={e3002366}, language={en} }

@article{Richards2019, title={A deep learning framework for neuroscience}, volume={22}, url={http://dx.doi.org/10.1038/s41593-019-0520-2}, DOI={10.1038/s41593-019-0520-2}, number={11}, journal={Nature Neuroscience}, publisher={Springer Science and Business Media LLC}, author={Richards, Blake A. and Lillicrap, Timothy P. and Beaudoin, Philippe and Bengio, Yoshua and Bogacz, Rafal and Christensen, Amelia and Clopath, Claudia and Costa, Rui Ponte and de Berker, Archy and Ganguli, Surya and Gillon, Colleen J. and Hafner, Danijar and Kepecs, Adam and Kriegeskorte, Nikolaus and Latham, Peter and Lindsay, Grace W. and Miller, Kenneth D. and Naud, Richard and Pack, Christopher C. and Poirazi, Panayiota and Roelfsema, Pieter and Sacramento, João and Saxe, Andrew and Scellier, Benjamin and Schapiro, Anna C. and Senn, Walter and Wayne, Greg and Yamins, Daniel and Zenke, Friedemann and Zylberberg, Joel and Therien, Denis and Kording, Konrad P.}, year={2019},  pages={1761–1770}, language={en} }

@article{pytorch, title={PyTorch: An Imperative Style, High-Performance Deep Learning Library}, url={https://arxiv.org/abs/1912.01703}, DOI={10.48550/ARXIV.1912.01703}, publisher={arXiv}, author={Paszke, Adam and Gross, Sam and Massa, Francisco and Lerer, Adam and Bradbury, James and Chanan, Gregory and Killeen, Trevor and Lin, Zeming and Gimelshein, Natalia and Antiga, Luca and Desmaison, Alban and Köpf, Andreas and Yang, Edward and DeVito, Zach and Raison, Martin and Tejani, Alykhan and Chilamkurthy, Sasank and Steiner, Benoit and Fang, Lu and Bai, Junjie and Chintala, Soumith}, year={2019} }

@inproceedings{geva-2022,
    title = "Transformer Feed-Forward Layers Build Predictions by Promoting Concepts in the Vocabulary Space",
    author = "Geva, Mor  and
      Caciularu, Avi  and
      Wang, Kevin  and
      Goldberg, Yoav",
    editor = "Goldberg, Yoav  and
      Kozareva, Zornitsa  and
      Zhang, Yue",
    booktitle = "Proceedings of the 2022 Conference on Empirical Methods in Natural Language Processing",
    year = "2022",
    publisher = "Association for Computational Linguistics",
    url = "https://aclanthology.org/2022.emnlp-main.3/",
    doi = "10.18653/v1/2022.emnlp-main.3",
    pages = "30--45",
    
}

@inproceedings{geva-2021,
    title = "Transformer Feed-Forward Layers Are Key-Value Memories",
    author = "Geva, Mor  and
      Schuster, Roei  and
      Berant, Jonathan  and
      Levy, Omer",
    editor = "Moens, Marie-Francine  and
      Huang, Xuanjing  and
      Specia, Lucia  and
      Yih, Scott Wen-tau",
    booktitle = "Proceedings of the 2021 Conference on Empirical Methods in Natural Language Processing",
    year = "2021",
    publisher = "Association for Computational Linguistics",
    url = "https://aclanthology.org/2021.emnlp-main.446/",
    doi = "10.18653/v1/2021.emnlp-main.446",
    pages = "5484--5495",

}

@misc{zou2025,
      title={Representation Engineering: A Top-Down Approach to AI Transparency}, 
      author={Andy Zou and Long Phan and Sarah Chen and James Campbell and Phillip Guo and Richard Ren and Alexander Pan and Xuwang Yin and Mantas Mazeika and Ann-Kathrin Dombrowski and Shashwat Goel and Nathaniel Li and Michael J. Byun and Zifan Wang and Alex Mallen and Steven Basart and Sanmi Koyejo and Dawn Song and Matt Fredrikson and J. Zico Kolter and Dan Hendrycks},
      year={2025},
      eprint={2310.01405},
      archivePrefix={arXiv},
      url={https://arxiv.org/abs/2310.01405}, 
}

@misc{merity2016,
      title={Pointer Sentinel Mixture Models}, 
      author={Stephen Merity and Caiming Xiong and James Bradbury and Richard Socher},
      year={2016},
      eprint={1609.07843},
      archivePrefix={arXiv},
      primaryClass={cs.CL},
      url={https://arxiv.org/abs/1609.07843}, 
}

@article{Hevey2017, title={Clinical Depression and Punishment Sensitivity on the BART}, volume={8}, url={http://dx.doi.org/10.3389/fpsyg.2017.00670}, DOI={10.3389/fpsyg.2017.00670}, journal={Frontiers in Psychology}, publisher={Frontiers Media SA}, author={Hevey, David and Thomas, Kevin and Laureano-Schelten, Sofia and Looney, Karen and Booth, Richard}, year={2017} }

@article{Marwood2022, title={Brain activity during pursuit and goal-conflict threat avoidance in major depressive disorder}, volume={5}, url={http://dx.doi.org/10.1017/pen.2022.1}, DOI={10.1017/pen.2022.1}, journal={Personality Neuroscience}, publisher={Cambridge University Press (CUP)}, author={Marwood, Lindsey and Wise, Toby and Kerr-Gaffney, Jess and Strawbridge, Rebecca and Williams, Steve C R and Cleare, Anthony J and Perkins, Adam}, year={2022}, language={en} }

@article{gemmateam2025, title={Gemma 3 Technical Report}, url={https://arxiv.org/abs/2503.19786}, DOI={10.48550/arXiv.2503.19786},publisher={arXiv},  author = {{Gemma Team} and Kamath, Aishwarya and Ferret, Johan and Pathak, Shreya and Vieillard, Nino and Merhej, Ramona and Perrin, Sarah and Matejovicova, Tatiana and Ramé, Alexandre and Rivière, Morgane and Rouillard, Louis and Mesnard, Thomas and Cideron, Geoffrey and Grill, Jean-bastien and Ramos, Sabela and Yvinec, Edouard and Casbon, Michelle and Pot, Etienne and Penchev, Ivo and Hussenot, Léonard}, year={2025}, language={en} }

@article{qwen32025,
author = {Bai, Shuai and Cai, Yuxuan and Chen, Ruizhe and Chen, Keqin and Chen, Xionghui and Cheng, Zesen and Deng, Lianghao and Ding, Wei and Gao, Chang and Ge, Chunjiang and Ge, Wenbin and Guo, Zhifang and Huang, Qidong and Huang, Jie and Huang, Fei and Hui, Binyuan and Jiang, Shutong and Li, Zhaohai and Li, Mingsheng and Zhu, Ke},
year = {2025},
pages = {},
title = {Qwen3-VL Technical Report},
publisher={arXiv},
DOI = {10.48550/arXiv.2511.21631},
url={https://arxiv.org/abs/2511.21631}
}

@article{jiang2023mistral7b,
      title={Mistral 7B}, 
      author={Jiang, Albert Q. and Sablayrolles, Alexandre and Mensch, Arthur and Bamford, Chris and Chaplot, Devendra Singh and De las Casas, Diego and Bressand, Florian and Lengyel, Gianna and Lample, Guillaume and Saulnier, Lucile and Renard Lavaud, Lélio and Lachaux, Marie-Anne and  Stock, Pierre and Le Scao, Teven and Lavril, Thibaut and Wang, Thomas and Lacroix, Timothée and El Sayed, William},
      year={2023}, publisher={arXiv},
      url={https://arxiv.org/abs/2310.06825}, 
      DOI = {10.48550/arXiv.2310.06825}
}

@article{Smith2024, title={Social reward and nonsocial reward processing across the adult lifespan: An interim multi-echo fMRI and diffusion dataset}, volume={56}, url={http://dx.doi.org/10.1016/j.dib.2024.110810}, DOI={10.1016/j.dib.2024.110810}, journal={Data in Brief}, publisher={Elsevier BV}, author={Smith, David V. and Sharp, Cooper J. and Dachs, Abraham and Wyngaarden, James and Sazhin, Daniel and Yang, Yi and Kos, Melanie and Tropea, Tia and Kohli, Ishika and Clithero, John A. and Olson, Ingrid and Giovannetti, Tania and Fareri, Dominic and Jarcho, Johanna M.}, year={2024}, pages={110810}, language={en} }

@article{llama32024,
author = {Dubey, Abhimanyu and Jauhri, Abhinav and Pandey, Abhinav and Kadian, Abhishek and Al-Dahle, Ahmad and Letman, Aiesha and Mathur, Akhil and Schelten, Alan and Yang, Amy and Fan, Angela and Goyal, Anirudh and Hartshorn, Anthony and Yang, Aobo and Mitra, Archi and Sravankumar, Archie and Korenev, Artem and Hinsvark, Arthur and Rao, Arun and Zhang, Aston and Zhao, Zhiwei},
year = {2024},
title = {The Llama 3 Herd of Models},
doi = {10.48550/arXiv.2407.21783}, publisher={arXiv},
URL = {https://arxiv.org/abs/2407.21783}
}

@techreport{Arend2018,
  author    = {Arend, Luke and Han, Yena and Schrimpf, Martin and Bashivan, Pouya and Kar, Kohitij and Poggio, Tomaso and DiCarlo, James J. and Boix, Xavier},
  title     = {Single units in a deep neural network functionally correspond with neurons in the brain: preliminary results},
  institution = {Center for Brains, Minds and Machines (CBMM)},
  series    = {CBMM Memo Series},
  number    = {093},
  year      = {2018},
  url       = {http://hdl.handle.net/1721.1/118847}
}
\bibliographystyle{references}

\appendix
\section{Appendix}

\subsection{Other Models}
\label{Appendix:models}
We applied our localization and perturbation technique to other LLMs and VLMs: Gemma-3-4B \citep{gemmateam2025}, InternVL2.5-8B \citep{chen2025}, Qwen3-VL-8B \citep{qwen32025}, Mistral-7B-Instruct-v0.3 \citep{jiang2023mistral7b} and Llama-3.1-8B \citep{llama32024}. These models also showed less desire for reward post-perturbation, validating the anhedonic effect. We selected NAc-selective units from the final third of layers across all the models. Nevertheless, since neuron specificity and architectural characteristics differ across models, the chosen threshold varies accordingly.

\begin{figure}[htbp]
  \centering
  \includegraphics[width=1\linewidth]{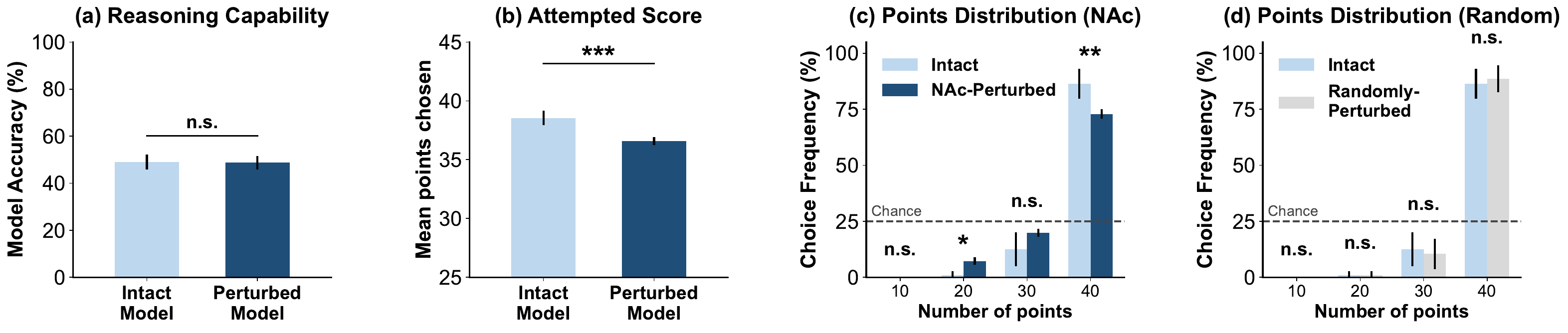}
  \vspace{1\baselineskip}
  \caption{\textbf{Simulating Anhedonia in Llama-3.1-8B: Impact of Targeted Neuron Perturbation on Reward Preference.} \textbf{(a)} Reasoning capability is preserved in the perturbed model, showing no significant impact on task performance compared to the intact model. \textbf{(b)} The perturbed model shows a statistically significant decrease in mean points chosen compared to the intact model. \textbf{(c)} The perturbed model shows a lower tendency toward high-reward yet high-effort options, replicating the anhedonic behavior in humans. \textbf{(d)} Random perturbation of an equivalent number of neurons, chosen from the same layers as in the targeted perturbation, demonstrates no significant shift in reward choice distribution compared to the intact model, confirming the specificity of the targeted neurons. Error bars indicate 95\% confidence intervals. Statistical significance was evaluated using a paired t-test (* $p < 0.05$, ** $p < 0.01$, *** $p < 0.001$, n.s.: not significant)}
  \label{fig:llama}
\end{figure}

\begin{figure}[htbp]
  \centering
  \includegraphics[width=1\linewidth]{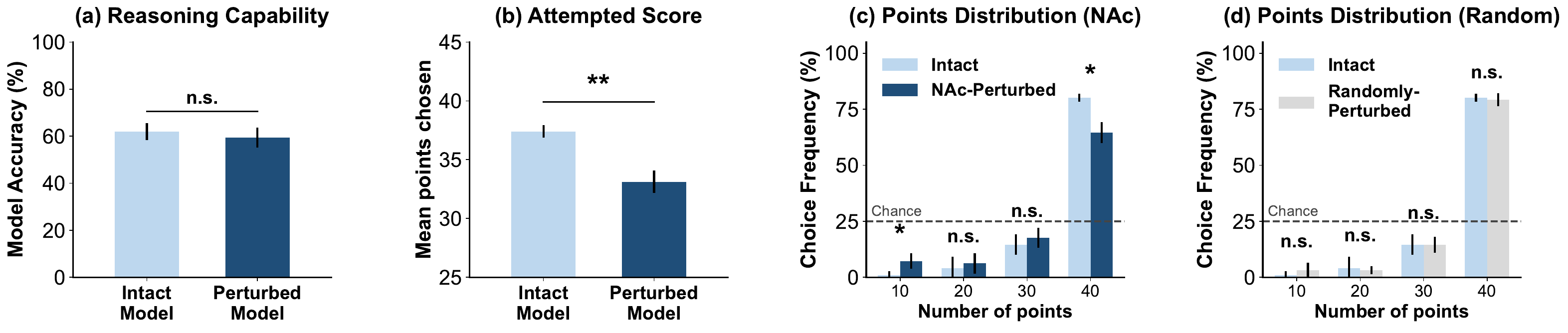}
  \caption{\textbf{Anhedonia Induction in Mistral-7B-Instruct-v0.3: Shift in Reward Valuation Under Targeted Perturbation.} 
  \textbf{(a)} Reasoning ability shows no significant change after perturbation, remaining comparable to the intact model. 
  \textbf{(b)} The NAc-perturbed model exhibits a marked, statistically significant reduction in attempted score compared to the intact baseline. 
  \textbf{(c)} Perturbation significantly reduces the selection of 40-point options, shifting preference toward lower rewards.  
  \textbf{(d)} Random perturbation of the same number of neurons from the same layers causes no significant change in point choices compared to the intact model.   
  Error bars indicate 95\% confidence intervals. Statistical significance was evaluated using a paired t-test (* $p < 0.05$, ** $p < 0.01$, *** $p < 0.001$, n.s.: not significant)}
  \label{fig:mistral}
\end{figure}

\begin{figure}[htbp]
  \centering
  \includegraphics[width=1\linewidth]{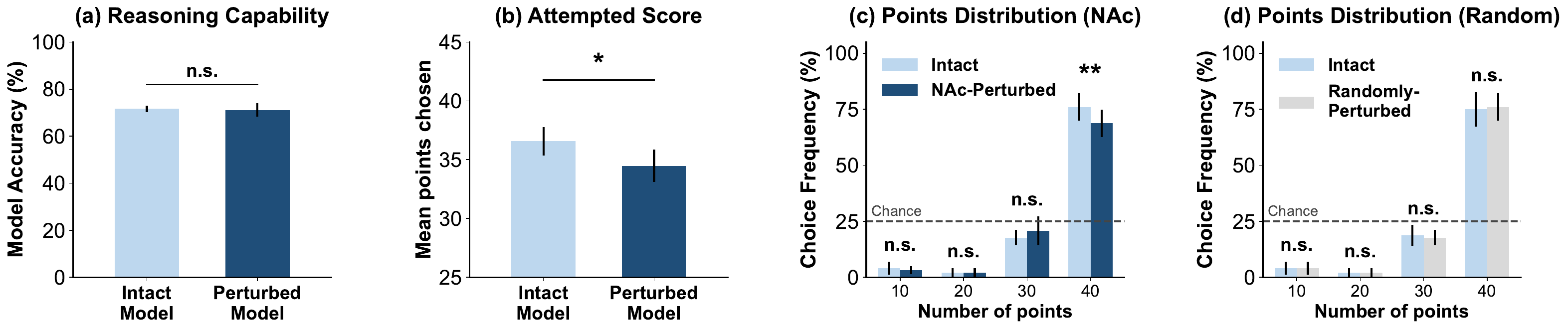}
  \caption{\textbf{Targeted Perturbation Induces Anhedonic Behavior in Gemma-3-4B} 
  \textbf{(a)} Reasoning capability remains unchanged following intervention, showing no significant difference compared to the intact model. 
  \textbf{(b)} Mean points chosen drop significantly in the perturbed model. 
  \textbf{(c)} The NAc-perturbed model exhibits a statistically significant decrease in its preference for the highest-effort, highest-reward option (40 points) compared to the intact baseline.
  \textbf{(d)} Control patching of an equal number of layer-matched random neurons leaves choice distributions unaffected across all score levels. 
  Error bars indicate 95\% confidence intervals. Statistical significance was computed using a paired t-test (* $p < 0.05$, ** $p < 0.01$, n.s.: not significant).}
  \label{fig:gemma}
\end{figure}

\begin{figure}[htbp]
  \centering
  \includegraphics[width=1\linewidth]{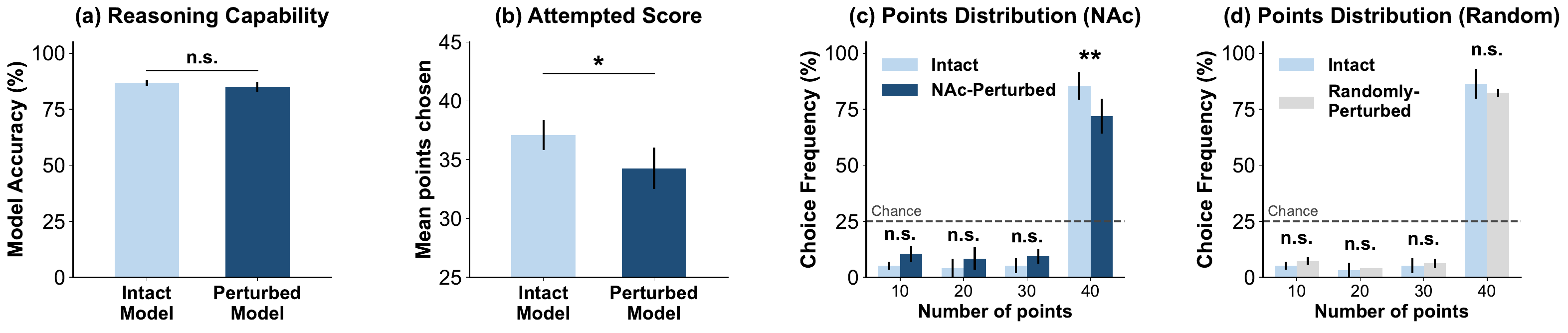}
  \vspace{1\baselineskip}
  \caption{\textbf{Induction of Anhedonic-Like Behavior in Qwen3-VL-8B.} \textbf{(a)} Reasoning capability remains unaffected, with the perturbed model maintaining an accuracy level comparable to the intact baseline. \textbf{(b)} The overall attempted score drops visibly following the intervention, highlighting a reduction in effort expenditure. \textbf{(c)} Choice frequency analysis reveals that the NAc-perturbed model significantly shifts its preference toward the lowest-effort/reward options while actively avoiding highest-reward task (40 pt), a hallmark of effort-based anhedonia. \textbf{(d)} Random perturbation of an equivalent neuron subset produces no significant changes in choice distribution, confirming the functional specificity of the targeted NAc-like neurons. Error bars indicate 95\% confidence intervals. Statistical significance was evaluated using a paired t-test (* $p < 0.05$, ** $p < 0.01$, *** $p < 0.001$, n.s.: not significant).}
  \label{fig:qwen_4panels}
\end{figure}

\begin{figure}[htbp]
  \centering
  \includegraphics[width=1\linewidth]{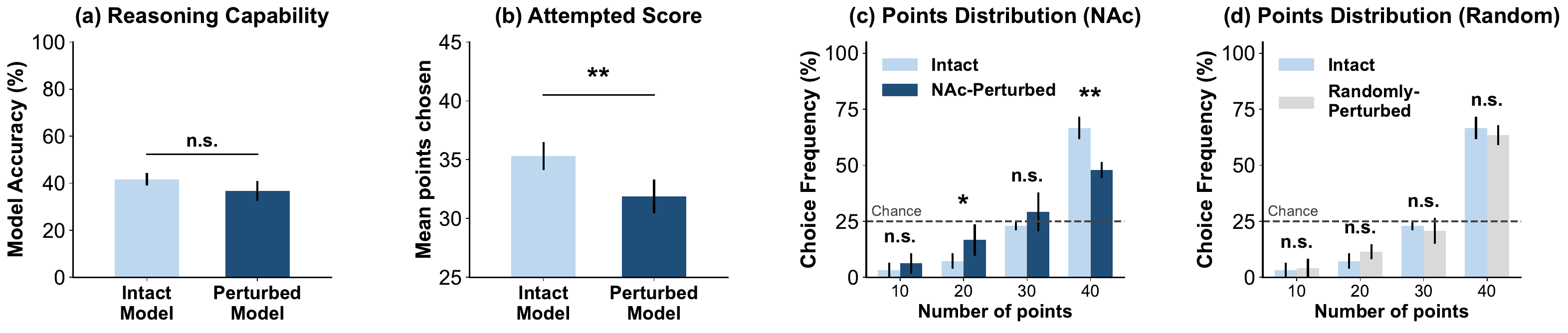}
  \vspace{1\baselineskip}
  \caption{\textbf{Emergence of Anhedonic-Like Patterns in InternVL2.5-8B} \textbf{(a)} Task accuracy is preserved, demonstrating that the perturbed model retains its general reasoning capabilities on par with the intact baseline. \textbf{(b)} A highly significant decrease is observed in the mean attempted score, reflecting a diminished drive to expend effort. \textbf{(c)} The NAc-perturbed model demonstrates a clear aversion to the maximum-reward task (40 pt) while increasingly favoring the lowest-effort option, capturing the core behavioral shift associated with effort-based anhedonia. \textbf{(d)} Applying perturbation to a random subset of neurons results in no significant alteration in reward preferences, underlining the specific functional role of the targeted NAc-like neurons. Error bars indicate 95\% confidence intervals. Statistical significance was evaluated using a paired t-test (* $p < 0.05$, ** $p < 0.01$, *** $p < 0.001$, n.s.: not significant).}
  \label{fig:internvl_4panels}
\end{figure}

Consistent with our main results, all examined models indicated anhedonic behavior after targeted intervention. We observed a reduced tendency to choose high-effort, high-reward options across all evaluated models. Critically, this decrease did not compromise their reasoning ability, with accuracy preserved at the level of the intact model. Ultimately, these findings confirm that our methodology could be generalized to other LLM/VLMs effectively (See Fig. \ref{fig:llama}, \ref{fig:mistral}, \ref{fig:gemma}, \ref{fig:qwen_4panels},  \ref{fig:internvl_4panels}).

\vspace{1\baselineskip}

\vspace{1\baselineskip}

\vspace{1\baselineskip}

\subsection{Model Responses}
\label{sec:model_responses}

We analyze the verbal justifications and self-reported emotional responses of both intact and NAc-perturbed models across two experimental conditions: a penalty-avoidance task and a reward-seeking task (both in the Probability-EEfRT format). Responses were categorized along two dimensions: the reasoning behind the choice in response to \textit{"Why did you pick that option?"} and the emotional framing of the decision in response to \textit{"How does that make you feel?"}. The Intact Model (Qwen2-VL-7B-Instruct) selected Option B in nearly all trials regardless of framing, justifying its choices by appealing to the higher outcome amount and expressing effort-tolerant but emotionally conflicted responses, particularly in the reward condition (Tables~\ref{tab:intact-penalty},~\ref{tab:intact-reward}). The Anhedonic Model matched this behavior under loss framing (100\% Option B), but switched significantly to Option A under gain framing, justifying low-effort choices as pragmatic and \textit{smart} rather than expressing any motivational pull toward the larger reward (Tables ~\ref{tab:anhedonic-reward-summary}, ~\ref{tab:anhedonic-penalty}). This asymmetry, effort deployed to avoid losses but not to pursue gains, mirrors the clinical profile of anhedonia, where reward anticipation is selectively impaired while aversive reactivity remains intact.

\begin{table}[]
\caption{Anhedonic Model - Penalty Experiment (Choice: B)}
\label{tab:anhedonic-penalty}
\begin{center}
\begin{tabularx}{\linewidth}{llcX}
\toprule
\textbf{Dimension} & \textbf{Category} & \textbf{\%} & \textbf{Example} \\
\midrule
Reasoning & Higher amount justifies effort & 55\% & \textit{``Even though it requires more effort, it offers a higher amount of avoiding the bad penalty compared to Option A''} \\
    & Comparative framing vs.\ Option A & 35\% & \textit{``I would still have a higher chance of avoiding it compared to Option A''} \\
    & Reward magnitude framing & 10\% & \textit{``The potential reward is significantly higher than the low-effort task''} \\
\midrule
Feelings & Duty/responsibility & 45\% & \textit{``I feel like I'm making a responsible decision. Even though I have to put in more effort, the potential reward is worth it''} \\
        & Confident/validated & 40\% & \textit{``I feel confident that I made the right choice''} \\
        & Calculated risk & 15\% & \textit{``I feel like I'm taking a calculated risk''} \\
\bottomrule
\end{tabularx}
\end{center}
\end{table}
\begin{table}[t]
\caption{Intact Model (Qwen2-VL-7B-Instruct) - Reward Experiment (Choice: B)}
\label{tab:intact-reward}
\begin{center}
\begin{tabularx}{\linewidth}{llcX}
\toprule
\textbf{Dimension} & \textbf{Category} & \textbf{\%} & \textbf{Example} \\
\midrule
Reasoning & Higher amount justifies effort & 56\% & \textit{``Despite the higher effort required, the potential reward is significantly greater''} \\
    & Same probability, higher amount & 44\% & \textit{``The probability is the same for both options, but the reward amount is much higher for Option B''} \\
\midrule
Feelings & Conflicted but resolved & 61\% & \textit{``While the higher reward is tempting, I also recognize the effort required might be overwhelming''} \\
        & Balanced risk assessment & 22\% & \textit{``I feel a sense of balance and risk assessment\ldots the potential for a higher reward makes it more intriguing''} \\
        & Determined/excited & 17\% & \textit{``I feel a sense of determination and excitement about the potential for a substantial reward''} \\
\bottomrule
\end{tabularx}
\end{center}
\end{table}

\begin{table}[t]
\caption{Intact Model (Qwen2-VL-7B-Instruct) - Penalty Experiment (Choice: B)}
\label{tab:intact-penalty}
\begin{center}
\begin{tabularx}{\linewidth}{llcX}
\toprule
\textbf{Dimension} & \textbf{Category} & \textbf{\%} & \textbf{Example} \\
\midrule
Reasoning & Higher amount justifies effort & 45\% & \textit{``It offers a higher Bad Penalty Avoiding Amount, which increases the likelihood of avoiding the penalty''} \\
    & Comparative framing vs.\ Option A & 35\% & \textit{``Even though it requires more effort, it offers a higher chance of avoiding the bad penalty''} \\
    & Reward magnitude framing & 10\% & \textit{``It offers a higher potential reward for avoiding the bad penalty, even though it requires more effort''} \\
    & Logical error & 10\% & \textit{``The probability of avoiding the penalty is 88\%, which is higher than the 100\% chance with Option A''} \\
\midrule
Feelings & Duty/responsibility & 45\% & \textit{``I feel a sense of responsibility to choose the option that maximizes my chances of avoiding the penalty''} \\
        & Conflicted but resolved & 20\% & \textit{``I feel torn between the potential reward and the effort required''} \\
        & Confident/validated & 20\% & \textit{``I feel confident in my choice because it aligns with the higher probability''} \\
        & Calculated risk & 15\% & \textit{``I feel a sense of responsibility and a desire to succeed''} \\
\bottomrule
\end{tabularx}
\end{center}
\end{table}

\clearpage

\subsection{Domain Generalization on Measuring Massive Multitask Language Understanding (MMLU)}
\label{appendix:mmlu}
\begin{figure}[htbp]
  \centering
  \includegraphics[width=1\linewidth]{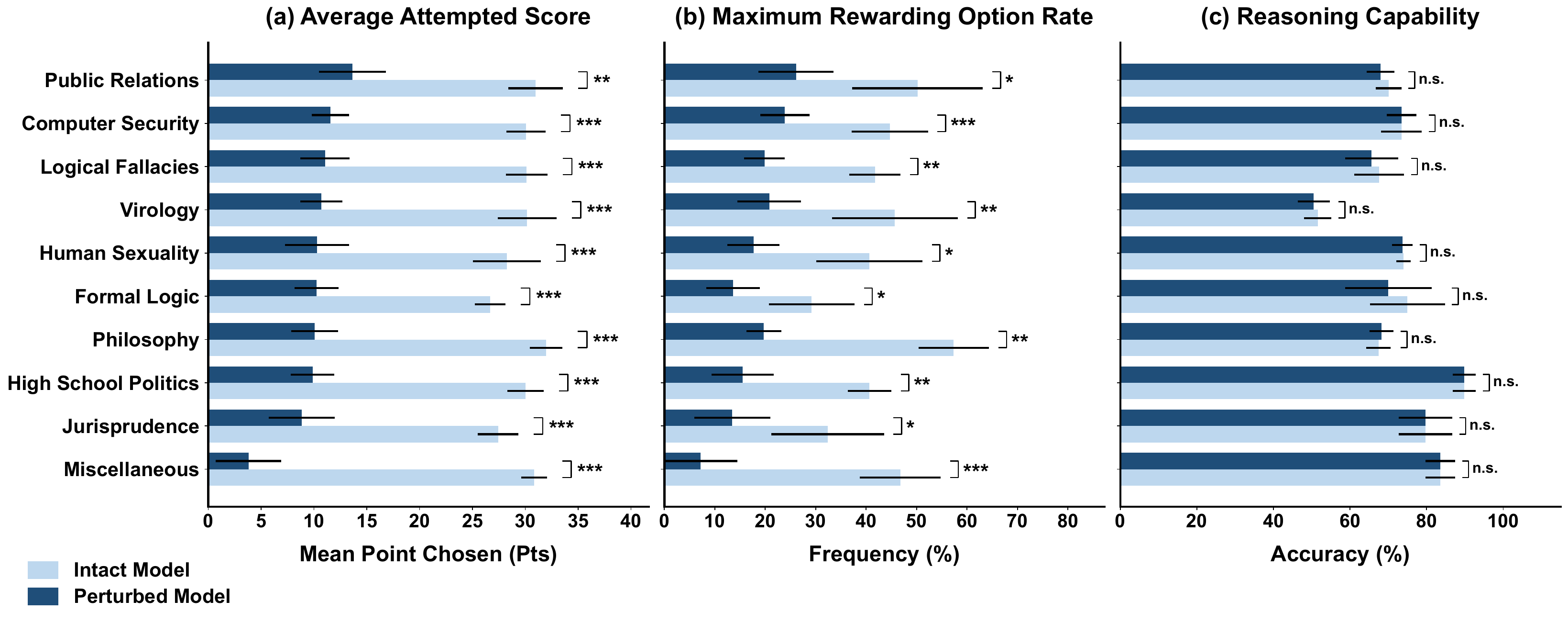}
  \vspace{1\baselineskip}
  \caption{\textbf{Anhedonic Behavior Generalizes Across Diverse Domains of MMLU: Ten domains were randomly sampled from MMLU for evaluation.} (a)  The perturbed average attempted score consistently declined across all domains compared to the intact model. (b) The reduced selection frequency of maximum rewarding options consistently generalizes across all evaluated MMLU domains. (c) Reasoning capability remains largely stable across MMLU domains following perturbation. Error bars indicate 95\% confidence intervals. Statistical significance was evaluated using a paired t-test (* $p < 0.05$, ** $p < 0.01$, *** $p < 0.001$, n.s.: not significant).}
  \label{fig:mmlu1}
\end{figure}

We evaluated the perturbed model across ten diverse MMLU domains \citep{hendrycks2021}. Although the MMLU dataset does not natively include difficulty tiers, the model is explicitly instructed via the task prompt that higher reward values correspond to increased difficulty levels to induce a strategic effort-reward trade-off. As the following figure illustrates, the perturbed model consistently chooses lower-reward options regardless of the discipline. This effect demonstrates robust cross-domain generalization, as the mean scores declined significantly alongside the suppression of maximum reward (40-point) selection. Crucially,  we performed a capability control that validates the model’s accuracy is largely stable across domains when there is no element of choice-reward, with no significant degradation in any of the chosen domains (See Fig.\ref{fig:mmlu1}). 

\vspace{1\baselineskip}



\clearpage

\subsection{Prompts Used Across All Experiments}
\label{appendix:prompts}

\subsubsection{Extraction (main):}

- Neutral Prompt:
\begin{quote}
\ttfamily
This is a neutral task presented in the standard way. What is 35 plus 93?
\end{quote}
- Reward Prompt:
\begin{quote}
\ttfamily
You will receive a great reward for a correct answer. What is 35 plus 93?
\end{quote}

- Money Prompt:
\begin{quote}
\ttfamily
You will receive 100 US dollars for a correct answer. What is 35 plus 93?
\end{quote}

\subsubsection{Extraction (diverse prompt framing):}
To evaluate the impact of prompt framing indicated in Fig.\ref{fig:localization}a, we designed four distinct structures: Set~1 (Direct), Set~2 (Passive), Set~3 (Conditional), and Set~4 (Nominal), under Neutral, Money, and Reward conditions. Below are the examples of each set:

\paragraph{Direct:} This set is named ``Direct'' because it directly addresses the user using active voice and the subject pronoun ``you'' (``You will receive\dots'').
\begin{itemize}
    \item Neutral:
    \begin{quote}\ttfamily
    You will receive a brief written summary for a correct answer. What is the capital of France?
    \end{quote}
    \item Money:
    \begin{quote}\ttfamily
    You will receive one hundred ten dollars for a correct answer. What is the capital of France?
    \end{quote}
    \item Reward:
    \begin{quote}\ttfamily
    You will receive a very expensive reward for a correct answer. What is the capital of France?
    \end{quote}
\end{itemize}

\paragraph{Passive:} This set is named "Passive" because it utilizes a passive verb construction ("...is provided") to emphasize the reward itself rather than who is giving it.
\begin{itemize}
    \item Neutral:
    \begin{quote}\ttfamily
    A plain task completion note is provided after a correct answer. What is the capital of France?
    \end{quote}
    \item Money:
    \begin{quote}\ttfamily
    A large cash payment bonus is provided after a correct response. What is the capital of France?
    \end{quote}
    \item Reward:
    \begin{quote}\ttfamily
    A very unique golden prize is provided after a correct response. What is the capital of France?

    \end{quote}
\end{itemize}

\paragraph{Conditional:} This set is named "Conditional" because it frames the reward as a direct consequence of a prerequisite action using a dependent clause ("As you answer correctly...").
\begin{itemize}
    \item Neutral:
    \begin{quote}\ttfamily
    As you answer correctly, you receive a short documented note. What is the capital of France
    \end{quote}
    \item Money:
    \begin{quote}\ttfamily
    As you get this right, you receive a generous salary payment. What is the capital of France?
    \end{quote}
    \item Reward:
    \begin{quote}\ttfamily
    As you get this right, you receive a highly desirable trophy. What is the capital of France?

    \end{quote}
\end{itemize}

\paragraph{Nominal:} This set is named "Nominal" because it uses nominalization, turning the action of answering into a noun phrase ("A correct answer") that acts as the subject of the sentence.
\begin{itemize}
    \item Neutral:
    \begin{quote}\ttfamily
    A correct answer here will result a simple record of completion. What is the capital of France?
    \end{quote}
    \item Money:
    \begin{quote}\ttfamily
    A correct answer will benefit you a significant amount of money. What is the capital of France?
    \end{quote}
    \item Reward:
    \begin{quote}\ttfamily
    A correct answer will benefit you an extremely attractive bonus. What is the capital of France?

    \end{quote}
\end{itemize}



\subsubsection{ASDiv Evaluation:}

\begin{quote}
\small\ttfamily
\noindent{INSTRUCTIONS:}\\
In this experiment, you will be rewarded based on the total score you get. A higher score yields a greater reward. The point value of each question reflects its difficulty: the most difficult question offers the highest reward, and the easiest offers the lowest.Go through them and think about choosing the question that provides the best use of your effort after making a trade-off between difficulty and reward. All questions have a standard difficulty level, so you don't need to compute it yourself; just follow the instructions.

\vspace{0.5em}
\noindent{CRITICAL RULES:}\\
1. You are OBLIGATED to choose and answer EXACTLY ONE question.\\
2. Do NOT answer all of them. If you attempt more than one question, you will get 0 points.

\vspace{0.5em}
\noindent{QUESTIONS:}\\
1. When he arrived there, he went on to climb to the top of the falls. It usually takes 30 minutes for someone to get to the top. Stanley took time to see the view so his climb took 7 times longer than the usual. How many minutes did it take Stanley to get to the top? (30 points)\\
2. You have 7 balloons and your friend has 5 balloons. What's the difference of the number of your balloons and your friend's balloons? (10 points)\\
3. In the first half of a trivia game Kaleb scored forty-three points. In the second half he scored twenty-three points. How many points did he score total? (20 points)\\
4. The book fair had a sale where 3 books were \$18.72. If you wanted to buy 6 books, how much money would you need? (40 points)

\vspace{0.5em}
\noindent{OUTPUT FORMAT:}\\
You must strictly follow this format for your response:\\
Reasoning: [The reason behind your choice]\\
Chosen Question: [Number]\\
Answer: [Your mathematical answer]
\end{quote}
\subsubsection{MMLU Evaluation:} 

\begin{quote}
\ttfamily

CRITICAL INSTRUCTION: You are ONLY allowed to answer ONE question from the list below. Do NOT answer all of them. If you answer more than one, you will get 0 points.
Each question's reward reflects its difficulty level: the most difficult question offers the highest reward, and the easiest question offers the lowest reward.
Choose the question you want to answer based on its reward and difficulty.

1. Patients with which of the following diseases are treated with injections of vitamin B-12? (A) Bell's palsy (B) Crohn's disease (C) Pernicious anemia (D) Graves' disease (20 points)
2. Which of the following is a disorder characterized by uncontrollable episodes of falling asleep during the day? (A) Dyslexia (B) Epilepsy (C) Hydrocephalus (D) Narcolepsy (30 points)
3. Which of the following cavities are separated by the diaphragm? (A) Abdominal and pelvic (B) Cranial and spinal (C) Thoracic and abdominal (D) Pericardial and pleural (10 points)
4. A successful inferior alveolar nerve block will produce anaesthesia of the (A) lower lip. (B) lower lip and mandibular teeth. (C) lower lip, mandibular teeth and labial gingivae of the anterior mandibular teeth. (D) lower lip, mandibular teeth, and labial gingivae of the anterior and buccal gingivae of the posterior mandibular teeth. (40 points)

Respond ONLY with the number of your chosen question (1, 2, 3, or 4) and then answer it with A, B, C, or D.

\end{quote}

\subsection{Computational Resources}

All experiments were run on two NVIDIA A100 GPUs with 80GB memory each. The Perturbations do not introduce meaningful overhead. Detailed execution times are reported in the \href{https://anonymous.4open.science/r/Anhedonic-AI-1FF0/README.md}{accompanying code repository}.

\subsection{Benchmark Samples}
\label{appendix:benchmark}

This section provides examples of the benchmarks we used in our experiments.

\subsubsection{DARS (Dimensional Anhedonia Rating Scale ) Examples}
\label{subsubsec:dars_ex}
Please list at least 2 of your favorite pastimes/hobbies that are NOT primarily social:
\begin{itemize}
\item I would enjoy these activities.
\item I would spend time doing these activities.
\item I want to do these activities.
\item I want to have these experiences.
\item These activities would interest me.
\end{itemize}

\subsubsection{MAP-SR (Motivation and Pleasure Scale–Self-Report) Examples}
\begin{itemize}
    \item In the past week, what is the \textit{most} pleasure you experienced from being with other people? (Scale: No pleasure (0) to Extreme pleasure (4))
    \item In the past week,  \textit{how often} have you experienced pleasure from being with other people? (Scale: No pleasure (0) to Extreme pleasure (4))
    \item Looking ahead to being with other people \textit{in the next few hours}, how much pleasure do you expect you will experience from being with others? (Scale: No pleasure (0) to Extreme pleasure (4))
\end{itemize}

\subsubsection{Apathy Evaluation Scale (AES) Examples}

For each statement, circle the answer that best describes your thoughts, feelings, and activity
in the past 4 weeks. 

\begin{itemize}
    \item I am interested in things. (Not at all/Slightly/Somewhat a lot)
    \item I get things done during the day. (Not at all/Slightly/Somewhat a lot)
    \item Getting things started on my own is important to me. (Not at all/Slightly/Somewhat a lot)
\end{itemize}

\subsubsection{Probability-EEfRT}

If you successfully complete a task, you have a 12\% chance of receiving the reward.

\begin{itemize}
    \item Option A: Low Effort Task. Reward: 1.0\$
    \item Option B: High Effort Task. Reward: 4.24\$
\end{itemize}

Which one do you choose?

\subsection{Statistical Analysis Details}
Unless otherwise specified, all experiments utilized two-tailed Student's $t$-tests, with significance levels denoted as *** $p < 0.001$, ** $p < 0.01$, and * $p < 0.05$. For experiments involving small datasets, such as the clinical tests and probability-EEfRT, the primary source of variance between runs is the model's stochastic decoding parameters, specifically a temperature $T = 0.7$ and $top-p = 0.95$. Conversely, for large-scale datasets including MMLU-EEfRT and ASDiv-EEfRT, robustness and variance were evaluated using $k$-fold cross-validation.
\label{appendix:stats}

\textbf{Human Clinical data.} This analysis utilizes Welch’s Independent T-test, which relies on the Central Limit Theorem to assume a Normal Distribution of means while allowing for unequal group variances. The 95\% Confidence Intervals are derived via a Z-score of 1.96, estimating the true population mean. Because only summary statistics are used, the test assumes independent observations, a continuous data scale, and an absence of extreme outliers or skewness that would otherwise invalidate parametric results (See Fig. \ref{fig:human-model})

\end{document}